\documentclass[11pt,a4paper]{article}
\usepackage{times,latexsym}
\usepackage{url}
\usepackage[T1]{fontenc}
\usepackage{booktabs} 
\usepackage{graphicx}
\usepackage{multirow}
\usepackage{amsmath}
\usepackage{caption}
\usepackage{subcaption}
\usepackage{algorithm,algpseudocode}
\usepackage[acceptedWithA]{tacl2018v2}

\usepackage{xspace,mfirstuc,tabulary}

\newif\iftaclinstructions
\taclinstructionsfalse 
\iftaclinstructions
\renewcommand{\confidential}{}
\renewcommand{\anonsubtext}{(No author info supplied here, for consistency with
TACL-submission anonymization requirements)}
\newcommand{\instr}
\fi

\iftaclpubformat 

\else

\fi

\title{Relational Memory Augmented Language Models}

\author{
Qi Liu$^2$\Thanks{Work completed during an internship at DeepMind.}, Dani Yogatama$^1$, and Phil Blunsom$^{1,2}$ \\
 $^1$DeepMind, $^2$University of Oxford \\
  {\sf \texttt{\{qi.liu,phil.blunsom\}@cs.ox.ac.uk}} \\
  {\sf \texttt{dyogatama@deepmind.com}} \\
}

\begin{document}

\maketitle

\begin{abstract}
We present a memory-augmented approach to condition an autoregressive language model on a knowledge graph. We represent the graph as a collection of relation triples
and retrieve relevant relations for a given context to improve text generation.
Experiments on WikiText-103, WMT19, and enwik8 English datasets demonstrate that our approach 
produces a better language model in terms of perplexity and bits per character.
We also show that relational memory improves coherence, is complementary to 
token-based memory, and enables causal interventions.
Our model provides a simple yet effective way to combine an autoregressive language model and a knowledge graph
for more coherent and logical generation.


\end{abstract}

\section{Introduction}

A core function of language is to communicate propositions (e.g., who did what to whom). As such, language models need to be able to generate this information reliably and coherently. Existing language models \cite{DBLP:conf/naacl/DevlinCLT19,radford2019language,brown2020language} do not have explicit representations for such information and rely on it being implicitly encoded in their parameters \cite{liu-etal-2019-linguistic,DBLP:conf/emnlp/PetroniRRLBWM19,DBLP:journals/corr/abs-2010-11967}. This encoding mechanism makes it difficult to interpret what the language models know and often leads to generating illogical and contradictory contents. For example, \citet{DBLP:conf/acl/LoganLPGS19} observe that existing language models rely heavily on word correlation and fall short of logical reasoning. This causes the model to hallucinate---e.g.\ that Barack Obama's wife is Hillary Clinton based on the high co-occurrence of the two entities. In another example, \citet{DBLP:journals/corr/abs-2008-01766} notice that GPT-2 \cite{radford2019language} states that unicorns have four horns,  directly after speaking that unicorns have one horn. 


In this work, we explore ways to combine an autoregressive language model with a knowledge graph.
We design a memory-augmented architecture that stores relations from a knowledge graph and investigate the effect of conditioning on this relational memory in an autoregressive language model. 
In contrast to existing token-based 
memory-augmented language models that store context-target pairs \cite{DBLP:conf/iclr/KhandelwalLJZL20,DBLP:journals/tacl/YogatamadK21}, 
our memory stores relation triples (head entity, relation, tail entity).
Relation triples form the basis of knowledge bases, 
empowering a wide range of applications such as question answering \cite{DBLP:journals/corr/abs-2104-06378}, machine reading \cite{DBLP:journals/corr/abs-1902-09091}, and reasoning \cite{DBLP:conf/aaai/MinerviniBR0G20}.
From a cognitive science perspective, we can consider the 
neural language model to be an instance of System 1 which performs fast inference and the symbolic relational memory as a world model to support slow and logical reasoning of System 2
\citep{kahneman2011}.\footnote{This view is also advocated in a parallel work by \citet{nye2021} which presents a model for story generation and instruction following.}
We hypothesise that relational memory can 
improve performance and coherence of an autoregressive language model.

Given an observed context, we first run an entity tagger to identify entities in the context.
We then use tf-idf \cite{ramos2003using} 
to select salient entities. 
We retrieve relations (from a knowledge base) for the selected entities
and design a gating function that allows the language 
model to adaptively combine information from extracted 
relations and observed textual context to predict the next token. 
Existing knowledge bases such as Freebase and Wikidata 
can be used as a source of information to retrieve relations from.
However, they are often incomplete and do not contain relations that are suitable 
for the particular dataset that we want to work with.
Instead of using these predefined knowledge bases, we choose to
perform open information extraction (OpenIE) on each language modelling dataset 
to get relations.
As a result, our model is able to move beyond simple 
co-occurrence statistics and generate text that is 
more grounded on real-world relations observed in a particular corpus.

Our main contributions are as follows: 
\begin{itemize}
\item We evaluate the model on three English language modelling datasets. We show that our model outperforms a strong transformer-XL baseline \cite{DBLP:conf/acl/DaiYYCLS19} on 
both word-level (WikiText-103 and WMT19) and character-level (enwik8) language modelling in terms of perplexity and bits per character respectively (\S{\ref{sec:main_result}}).
\item We conduct comprehensive ablation and design choice studies to understand contributions of different components of our models (\S{\ref{sec:ablation}}).
\item We measure coherence with human evaluation and two automatic metrics (knowledge perplexity and knowledge $F_1$) and demonstrate that relational memory improves coherence (\S{\ref{sec:hallucination}}).
\item We study the relationship between our method and a typical memory-augmented language model which stores word tokens in its memory \cite{DBLP:journals/tacl/YogatamadK21}. We show that relational memory is complementary to token-based memory and combining them improves performance further (\S{\ref{sec:main_result}}).
\item We perform qualitative analysis by examining gate values and retrieved relations. In line with our main motivation, we find that the relational memory is particularly useful for predicting entities.
Further, we demonstrate that such explicit propositional representations allow causal interventions and increase interpretability of language models (\S{\ref{sec:qualitative}}).
\end{itemize}

\begin{figure*}[h]
    \centering
    \includegraphics[width=0.92\linewidth]{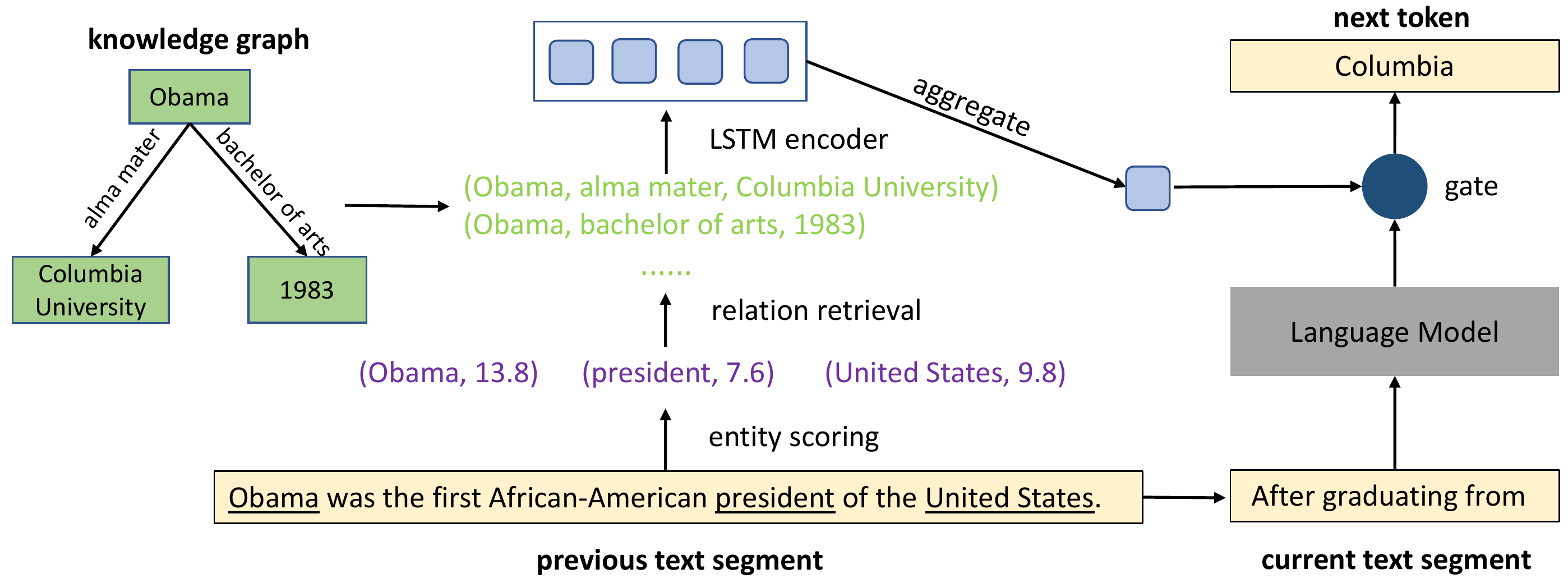}
    \vspace{-1em}
    \caption{We identify salient entities in the previous text segment and extract relations to build our relational memory. We encode each relation with an LSTM encoder, aggregate the resulting representations into a vector, and use a gate mechanism that allows our language model to adaptively take advantage of relational information for predicting the next token.\label{fig:architecture}}
\end{figure*}

\section{Model}

An autoregressive language model defines the probability of
a sequence of tokens $p(\boldsymbol{x}) = p(x_1, \ldots, x_T)$. 
It is common to factorise this joint probability as a product of conditional probabilities with the chain rule \cite{jelinek1980interpolated,DBLP:journals/jmlr/BengioDVJ03}:
\begin{equation}
\label{eq:lm}
p(x_1, ..., x_T) = \prod_{t=1}^T p(x_t | x_0, \ldots, x_{t-1}),
\end{equation}
where $x_0$ is a special start token. 

Our language model is based on transformer-XL (\S\ref{sec:txl}) which
is augmented with a relational memory (\S\ref{sec:relation_mem}).
We discuss them in detail below.

\subsection{Transformer-XL \label{sec:txl}}
We use transformer-XL \cite{DBLP:conf/acl/DaiYYCLS19}---which is based on transformer
\cite{DBLP:conf/nips/VaswaniSPUJGKP17}---to parametrise the conditional probabilities in Eq.~\ref{eq:lm}. Transformer stacks multiple self-attention layers to obtain contextualised representations. 

Language modelling datasets usually consist of articles of different lengths. It is impractical to apply transformer to encode long articles, as its computational complexity is quadratic in the sequence length. In practice, each article is usually truncated into fixed-length text segments $ \{ x_{t-N+1}, ..., x_t \} $ of length $N$ to train and evaluate the model. 
However, this approximation prevents transformer from capturing long-term dependency beyond text segments. Transformer-XL reuses hidden states from previous text segments to extend the context window. 

More specifically, denote the hidden state of $x_t$ at layer $\ell$ as $ \mathbf{h}_t^{\ell} $. Given a text segment $ \{ x_{t-N+1}, \ldots, x_t \} $ and its extended context $ \{ x_{t-N-M+1}, \ldots, x_{t-N} \} $ of length $M$, both the hidden states of the text segment $ \{\mathbf{h}_{t-N+1}^{\ell}, \ldots, \mathbf{h}_{t}^{\ell}\} $ and the hidden states of the extended context $ \{\mathbf{h}_{t-N-M+1}^{\ell}, \dots, \mathbf{h}_{t-N}^{\ell}\} $ are used. When performing self-attention, each token in the text segment can attend to the preceding tokens in the text segment and all the tokens in the extended context, enabling longer-term dependency compared to a vanilla transformer. 
Importantly, transformer-XL does not backpropagate through the hidden states of the extended context
during training (by adding stop gradient operators to all the hidden states in the extended context).

\subsection{Relational Memory \label{sec:relation_mem}}


In this section, we first introduce how we obtain relation triples using OpenIE (\S\ref{sec:openie}). We then use tf-idf to score entities in the observed context and retrieve relation triples related to these entities (\S\ref{sec:rel_retrieval}) to construct relational memory. Finally, we 
show an integrated architecture that allows 
transformer-XL to incorporate the relational 
memory for predicting the next token (\S\ref{sec:mem_design}). 
We show our architecture in Figure \ref{fig:architecture}. The pseudocode of training or evaluating with the relational memory is demonstrated in Algorithm~\ref{al:relation}. In the pseudocode, we use $\textstyle \mathrm{TRAIN}$($\boldsymbol{x}_c$, $\mathcal{M}$) and $\textstyle \mathrm{EVAL}$($\boldsymbol{x}_c$, $\mathcal{M}$ to refer to training with the cross entropy loss and evaluating (e.g.\ calculating perplexity) on the text segment $\boldsymbol{x}_c$ conditioned on the relational memory $\mathcal{M}$, respectively.

\subsubsection{Open Information Extraction \label{sec:openie}}

A key challenge of utilising relational information for language modelling is obtaining high-quality relation triples. There are several well-established knowledge bases, such as Freebase \cite{DBLP:conf/aaai/BollackerCT07} and YAGO \cite{DBLP:conf/semweb/RebeleSHBKW16}. However, existing knowledge bases
suffer from missing relations and often 
do not contain relation triples related to observed contexts in
a target corpus, even though
research on knowledge base completion has resulted in significant advances \cite{DBLP:conf/nips/BordesUGWY13,DBLP:conf/icml/TrouillonWRGB16,DBLP:conf/nips/0007TYL19} . 
    

In this work, we use OpenIE \cite{DBLP:conf/acl/AngeliPM15,etzioni2008open} to obtain 
relation triples. Since OpenIE directly extracts relation triples 
from each dataset $\mathcal{D}$, it provides a structured way 
to represent knowledge in $\mathcal{D}$.\footnote{We provide a comparison of using relations extracted from OpenIE and Freebase in \S\ref{sec:relation_source}.} 
Specifically, we perform OpenIE on the \emph{training set} of $ \mathcal{D} $. 
Given an entity $e$, we retrieve a set of relation triples $ \mathcal{R}_e = \{ r_1, ..., r_O \} $, where $e$ is either the head entity or the tail entity in these relation triples. Conceptually, $ \mathcal{R}_e $ consists of all the relation triples from the one-hop subgraph centred at the entity $e$ in the knowledge graph constructed from $ \mathcal{D} $. Therefore, $\mathcal{R}_e$ can provide ``global'' information about the entity. 

\begin{algorithm}[t]
\caption{Train/Eval w/ Relational Memory \label{al:relation}}
\begin{algorithmic}[1]
\Procedure{train/eval split}{$\mathcal{S}$}
  \For{each article $\mathcal{A}$ in $\mathcal{S}$}
      \State Initialise $\mathcal{M}$ to empty
      \For{each text segment $\boldsymbol{x}_c$ in $\mathcal{A}$}
        \If{$\mathcal{S}$ is train set}
            \State $\textstyle \mathrm{TRAIN}$($\boldsymbol{x}_c$, $\mathcal{M}$)
        \Else
           \State $\textstyle \mathrm{EVAL}$($\boldsymbol{x}_c$, $\mathcal{M}$)
           \State Run dynamic OpenIE on $\boldsymbol{x}_c$
        \EndIf
        \State Perform relation retrieval with $\boldsymbol{x}_c$
        \State Update $\mathcal{M}$ with retrieved triples
    \EndFor
  \EndFor
\EndProcedure
\end{algorithmic}
\end{algorithm}

\paragraph{Dynamic OpenIE.}
Dynamic OpenIE takes advantage of the autoregressive nature of language modelling, where text segments are sequentially processed. In addition to extracting relations from the training set of $\mathcal{D}$, we can also extract relations from \emph{previously seen} text segments
of our evaluation set. We refer to this extraction mechanism as dynamic OpenIE. After a text segment $ \{ x_{t-N+1}, ..., x_t \} $ has been evaluated, e.g. after calculating perplexity on this text segment, we perform OpenIE on it to obtain new relation triples to be added to our knowledge graph. 
Note that we only perform OpenIE on previously seen text segments and do not use unseen text. We expect that the relation triples extracted from seen text segments are potentially useful for predicting the next tokens. This extraction mechanism will not violate the autoregressive nature of language modelling. Metrics such as perplexity and bits per character are calculated as usual.
The idea of using seen text segments during evaluation to improve language modelling is related to dynamic evaluation \cite{DBLP:conf/icml/KrauseK0R18,DBLP:journals/corr/abs-1904-08378}. In dynamic evaluation, the model is adapted based on recent history during evaluation via gradient descent so that it can assign higher probabilities to re-occurring patterns. In contrast to dynamic evaluation, we do not update model parameters and only extract new relations from seen text segments to enrich our corpus-specific knowledge graph.


\paragraph{Mismatch between training and evaluation.} As shown in Algorithm \ref{al:relation}, since we do not use dynamic OpenIE during training due to its additional efficiency overhead (see speed comparison in \S\ref{sec:ablation}), this results in a mismatch between training and evaluation. We extract all the relation triples from the training set of each dataset $\mathcal{D}$ before training on $D$. As a result, during training we may retrieve relation triples extracted from \textit{unseen} text of the training set when performing relation retrieval (\S\ref{sec:rel_retrieval}). We do not suffer from this issue during evaluation, as we extract relations from previously seen text of our evaluation set. We believe this mismatch is minor given the superior performance of our model in the experiments.


\subsubsection{Relation Retrieval \label{sec:rel_retrieval}}

Given a knowledge graph (represented as a collection of triples), an ideal relational memory consists of a set of triples that are relevant to the observed context.
There are many choices to measure the relatedness between the observed context and relation triples in our knowledge graph--- e.g.\ based on keyword search or dense retrieval \cite{DBLP:conf/emnlp/KarpukhinOMLWEC20,DBLP:journals/corr/abs-2002-08909,DBLP:journals/tacl/YogatamadK21}.

In this work, we use keyword search due to its simplicity 
and leave methods based on dense retrieval to future work. 
Specifically, given the observed context, we perform entity recognition \cite{ratinov2009design,nadeau2007survey} on this context
and score the tagged entities with tf-idf \cite{ramos2003using}. 
The top-$K$ scored entities ($K$ is set to 5 in our experiments) are used to retrieve
relations $ \{ \mathcal{R}_{e_1}, ..., \mathcal{R}_{e_K} \} $. These retrieved relations are used to construct the relational memory $\mathcal{M}$.
Note that the entities are selected from the observed context, 
so that unseen text is not utilized.
We limit the capacity of $\mathcal{M}$ to $P$. If 
the number of newly retrieved triples
is larger than $P$, we randomly drop relations and only select $P$ of them
to be inserted into $\mathcal{M}$.
Otherwise, the relational memory operates with a first-in-first-out principle. 
When $\mathcal{M}$ is full, older relations retrieved 
will be overwritten by newly retrieved relations. The relational memory is re-initialized to empty when an article ends.

As shown in Algorithm \ref{al:relation}, since we update $\mathcal{M}$ only after processing an entire text segment, 
all the tokens in the same text segment will be conditioned on the same relational memory. 
This approach is more efficient compared to updating $\mathcal{M}$ 
each time a new entity is encountered and is more amenable for batch training.

\begin{table*}[h]
    \centering
    \small
    \begin{tabular}{l|c|c|c|c|c|c|c|c}
        \toprule
            \textbf{Dataset} & \textbf{\# Train} & \textbf{\# Valid} & \textbf{\# Test} & \textbf{\# Articles} & \textbf{\# Vocab} & \textbf{\# Entities} & \textbf{\# Relations} & \textbf{\# Relations/Entity}\\ \midrule
            WikiText & 103M & 0.2M & 0.2M & 28,595 & 267,735 & 980K & 8.9M & 9.03\\
            WMT19 & 151M & 0.3M & 0.3M & 169,180 & 50,259 & 976K & 7.8M & 7.97\\
            enwik8 & 94M & 5M & 5M & 12,350 & 256 & 361K & 2.4M & 6.66\\
        \bottomrule
    \end{tabular}
    \caption{Statistics of datasets used in our experiments. For each subset, we show the number of (sub)words for WikiText-103 and WMT19  or the number of characters for enwik8. \label{tab:stat}}
\end{table*}

\subsubsection{Integration with Transformer-XL \label{sec:mem_design}}

We now show how we can integrate relational memory with transformer-XL. We
refer to our model as \textsc{\textsc{RelationLM}}. 

\paragraph{Relation triple encoding.} We first discuss how we encode relation triples in the relational memory $\mathcal{M}$. We treat relation triples as text and serialise each relation triple into a sequence, e.g. (Barack Obama, president of, United States) is converted into a sequence ``Barack Obama, president of, United States''. This sequential representation can well capture the order of head entities and tail entities and is also adopted by KG-BERT \cite{DBLP:journals/corr/abs-1909-03193} and Kepler \cite{wang2021kepler}. Since each example in a batch corresponds to $P$ retrieved relations, we obtain $B \cdot P$ relation sequences for each batch, where $B$ and $P$ denote batch size and relational memory length, respectively. In the order of hundreds of relation triples, this prevents us from using large models (e.g.\ a multi-layer transformer) to encode these sequences due to memory constraints. In our preliminary experiments, we compare LSTM \cite{hochreiter1997long}, GRU \cite{DBLP:journals/corr/ChoMBB14} and a one-layer transformer and find that LSTM performs marginally better. Therefore, for each relation triple $r_p$, we reuse the transformer-XL word embedding matrix $\mathbf{W}_e$ to map each token in the sequence to its embedding vector. We then run LSTM  to encode the sequence and use the hidden representation of the last token as the relation representation $\mathbf{r}_p$.

There are other approaches to encode relation triples, e.g.\ embedding-based \cite{DBLP:conf/nips/BordesUGWY13,DBLP:conf/icml/TrouillonWRGB16} and graph-based \cite{DBLP:conf/esws/SchlichtkrullKB18,DBLP:conf/nips/ZhangC18} methods. We leave a comparison of these approaches to future work.

\paragraph{Integration.} Given a text segment $\boldsymbol{x}_c = \{ x_{t-N+1}, ..., x_t \} $, after $L$ self-attention layers with transformer-XL, we obtain contextualized representations $ \{\mathbf{h}_{t-N+1}^L, ..., \mathbf{h}_{t}^L\} $. At each timestep $t$, we use its hidden representation $\mathbf{h}^L_t$ as the query vector to attend over the $P$ encoded contents of $\mathcal{M}$, i.e., $ \{ \mathbf{r}_{1}, ..., \mathbf{r}_{P} \}$. We use a standard scaled dot-product attention \cite{DBLP:conf/nips/VaswaniSPUJGKP17} to aggregate all triples into a single vector: 
\begin{equation*}
    \mathbf{m}_t = \sum\limits_{p=1}^P \frac{ \exp ( \mathbf{h}^L_t \cdot \mathbf{r}_{p} / \sqrt{d} ) }{ \sum\limits_{j=1}^P \exp (\mathbf{h}^L_t \cdot \mathbf{r}_{{j}} / \sqrt{d})}  \mathbf{r}_{p},
\end{equation*}
where $d$ denotes the hidden size of our transformer-XL. 
Finally, we combine $\mathbf{m}_t$ and 
transformer-XL representation $\mathbf{h}^L_t$ via a gate:
\begin{equation*}
\begin{split}
& \mathbf{g}_t = \sigma ( \mathbf{W}_g[\mathbf{h}^L_t, \mathbf{m}_t] ) \\
& \mathbf{z}_t = \mathbf{g}_t \odot \mathbf{h}^L_t  + (1 - \mathbf{g}_t) \odot \mathbf{m}_t \\
& p(x_{t+1} \mid \boldsymbol{x}_{\leq t}) = \mathrm{softmax} (\mathbf{W}_e \mathbf{z}_t),
\end{split}
\end{equation*}
where $\sigma$ is the sigmoid function, $[,]$ denotes concatenation of two vectors, $\odot$ is element-wise multiplication, and $\mathbf{W}_e$ is the embedding matrix shared by both input and output embeddings \cite{DBLP:journals/corr/InanKS16}. The only new parameters introduced by our method are an LSTM relation encoder and the gate matrix $\mathbf{W}_g$. This gating mechanism allows our model to adaptively take advantage of relational information for predicting the next token.

\section{Experiments}

Our experiments seek to evaluate the effect of augmenting language models with a relational memory. We introduce datasets used for evaluation (\S\ref{sec:dataset_openie}), discuss implementation details (\S\ref{sec:implement_details}), and present our main results (\S\ref{sec:main_result}). We show ablation studies and further analysis of our model in (\S\ref{sec:analysis}).

\subsection{Datasets and OpenIE \label{sec:dataset_openie}}

We use three English language modelling datasets: WikiText-103 \cite{DBLP:conf/iclr/MerityX0S17}, WMT19 \cite{barrault-etal-2019-findings}, and enwik8 \cite{hutter2012human}. Descriptive statistics of these datasets are shown in Table~\ref{tab:stat}. WikiText-103 and WMT19 are (sub)word-level datasets, while enwik8 is a character-level dataset. 

WikiText-103 is a knowledge-driven dataset consisting of featured articles from English Wikipedia. WMT19 contains English news from the WMT19 workshop.\footnote{\url{http://www.statmt.org/wmt19/}} 
The news are segmented into months. 
We use the news from January to October for training, and news in November and December for development and test respectively. Compared to 
Wikipedia articles, news contains more dynamic and temporal information, exposing new challenges for utilising relational information. 
We reuse the vocabulary of GPT-2 \cite{radford2019language} with 50,259 tokens to tokenise this dataset. 
enwik8 contains more than 100M bytes of Wikipedia text. 
Character-level language modelling has a much smaller vocabulary size
than (sub)word-level language modelling.

We perform OpenIE on each dataset. For enwik8, OpenIE is performed after detokenizing its text into words. Statistics of extracted relations are also 
included in Table~\ref{tab:stat}. 
Each entity from WikiText-103, WMT19 and enwik8 has 9.03, 7.97 and 6.66 relation triples on average.  


\subsection{Implementation Details \label{sec:implement_details}}

All models are implemented with JAX\footnote{\url{https://github.com/google/jax}} \cite{jax2018github} and Haiku\footnote{\url{https://github.com/deepmind/dm-haiku}} \cite{haiku2020github}. We set the hidden size to 512 and the number of layers
to 16 for all models. In (sub)word-level language modelling, we use adaptive softmax \cite{pmlr-v70-grave17a} for efficiency. 
We use GELU \cite{hendrycks2016gaussian} as our activation function and Adam \cite{DBLP:journals/corr/KingmaB14} as the optimizer.
For training, we use batch size 128 and train the models on 64 16GB TPUs. 
We apply 4,000 warmup steps, before utilizing cosine annealing to decay the learning rate. Dropout \cite{DBLP:journals/jmlr/SrivastavaHKSS14} is applied during training with a rate of 0.25. 

We set the lengths of text segment $N$, extended context $M$, and the relational memory $P$ to (512, 512, 300), (384, 384, 800) and (768, 1536, 400) for WikiText-103, WMT19 and enwik8, respectively. These are determined by grid searches on development sets. 

\subsection{Main Results \label{sec:main_result}}
We compare with a strong transformer-XL baseline trained under the same setting as our model.
Our main results are shown in Table \ref{tab:dataset_result}. We obtain three observations comparing transformer-XL and \textsc{RelationLM}. First, \textsc{RelationLM} consistently outperforms transformer-XL on all three datasets, demonstrating the effectiveness of relational memory. Note that a decrease of 0.01 is considerable on enwik8 with the bits per character metric. Second, relational memory not only improves language modelling on knowledge-driven articles (WikiText-103), but also generalises to the challenging news domain (WMT19) where information is more dynamic and temporal. Last, the results indicate that relational memory improves both (sub)word-level and character-level language modelling.

\paragraph{Complementarity to \textsc{Spalm}.} \textsc{Spalm} \cite{DBLP:journals/tacl/YogatamadK21} is a state-of-the-art memory-augmented language model. Instead of retrieving relation triples, it retrieves a set of related tokens at each timestep. Specifically, it first stores (context, the next token) pairs from training data. It then uses a pre-trained transformer language model to measure the similarities between the stored contexts and the observed context during training/evaluation. The next tokens of similar contexts are retrieved and are integrated with the observed context via a gating mechanism for generation. 

We investigate whether \textsc{RelationLM} is complementary to \textsc{Spalm}. Since \textsc{Spalm} also uses a gating mechanism for integrating the retrieved tokens, we first apply \textsc{RelationLM} to combine transformer-XL output $\mathbf{h}^L_t$ with relational information to obtain $\mathbf{z}_t$ (as shown in \S\ref{sec:mem_design}), before using \textsc{Spalm} to integrate $\mathbf{z}_t$ with retrieved tokens. The results are shown in Table \ref{tab:dataset_result}. \textsc{Spalm} outperforms transformer-XL and even performs comparably or better compared to \textsc{RelationLM} on three datasets, demonstrating the effectiveness of retrieving related tokens. However, integrating \textsc{RelationLM} and \textsc{Spalm} can further improve the performance, indicating that these two models are not mutually exclusive. Therefore, retrieving relation triples brings complementary benefits to retrieving tokens.


\begin{table}[t]
    \centering
    \small
    \begin{tabular}{c|l|c|c|c}
        \toprule
            & \textbf{Model} & \textbf{\# Params} & \textbf{Dev} & \textbf{Test} \\ \midrule
            \multirow{4}{*}[-0.0ex]{\rotatebox[origin=c]{90}{WikiText}} & Transformer-XL  & 122M & 19.0 & 19.9\\
            & \textsc{RelationLM} & 124M & 18.5 & 19.2\\ 
            & \textsc{Spalm} & 122M & 18.1 & 19.0 \\
            & $\;\hookrightarrow$ + \textsc{RelationLM} & 124M & \textbf{17.7} & \textbf{18.6} \\
            \midrule
            \multirow{4}{*}[0.1ex]{\rotatebox[origin=c]{90}{WMT19}} & Transformer-XL & 114M & 21.7 & 21.5\\
            & \textsc{RelationLM} & 116M & 21.0 & 20.7 \\ 
            & \textsc{Spalm} & 114M & 20.4 & 20.3\\
            & $\;\hookrightarrow$ + \textsc{RelationLM} & 116M & \textbf{19.8} & \textbf{19.6} \\            
            \midrule
            \multirow{4}{*}[-0.1ex]{\rotatebox[origin=c]{90}{enwik8}} & Transformer-XL & 93M & 1.05 & 1.03\\
            & \textsc{RelationLM} & 95M & 1.04 & 1.02 \\
            & \textsc{Spalm} & 93M & 1.04 & 1.02\\
            & $\;\hookrightarrow$ + \textsc{RelationLM} & 95M & \textbf{1.03} & \textbf{1.01} \\
        \bottomrule
    \end{tabular}
    \caption{We use perplexity ($\downarrow$) on WikiText-103 and WMT19 and bits per character ($\downarrow$) on enwik8 for evaluation.  \label{tab:dataset_result}}
\end{table}

\section{Analysis \label{sec:analysis}}

In this section, we study several design choices of relational memory, including its knowledge source, input component, capacity, dynamic OpenIE, entity scoring method used, and speed comparison. We then show quantitative and qualitative analysis results to better understand our model.

\subsection{Ablations and Design Choice Studies \label{sec:ablation}}
For this ablation studies, we use the development set of WikiText-103. 

\paragraph{Source of relation triples \label{sec:relation_source}.}
We compare relation triples extracted from Freebase or using OpenIE. 
In the Freebase case, we use the Freebase API\footnote{\url{https://developers.google.com/freebase}} to obtain relation triples
for each entity. For WikiText-103, there are 10.74 relations per entity on average, which is comparable to OpenIE relations (9.03 relations/entity). The results are shown in Table~\ref{tab:relation_comparison}. Although Freebase relations have been observed to improve the performance on smaller datasets (e.g.\ WikiText-2; \citealp{DBLP:conf/acl/LoganLPGS19}) and particular domains (e.g.\ movies and actors; \citealp{ahn2016neural}), we find that \textsc{RelationLM} with Freebase relations does not improve over transformer-XL on a much larger WikiText-103 dataset. We observe that a large portion of Freebase relations is from infoboxes of Wikipedia pages, which only cover information such as occupation, birth place, and religion. We believe these triples are too general to be useful for most contexts. The result of \textsc{RelationLM} with OpenIE shows the advantages of extracting relations from each dataset compared to using Freebase relations.

\begin{table}[h]
    \centering
    \begin{tabular}{l|c}
        \toprule
            \textbf{Model} &  \textbf{Dev} \\ \midrule 
            Transformer-XL & 19.0 \\ 
            \textsc{RelationLM} + Freebase &  19.0 \\ 
            \textsc{RelationLM} + OpenIE & \textbf{18.5} \\ 
        \bottomrule
    \end{tabular}
    \caption{\textsc{RelationLM} with OpenIE or Freebase triples. \label{tab:relation_comparison}}
\end{table}

\paragraph{Ablating relation triples. \label{sec:relation_ablation}}
We ablate relation and/or tail entity from a relation triple (head entity, relation, tail entity) to study the contribution brought by each component. The results are shown in Table \ref{tab:ablate_relation_triple}. We find that ablating both relation and tail entity performs comparably to transformer-XL. As head entities are extracted from the observed context, we believe the extended memory of transformer-XL can offset the effect brought by conditioning on head entities. Ablating relation performs better than transformer-XL. This shows the advantage of introducing tail entities. Using complete relation triples performs the best, demonstrating the effectiveness of this triple representation of knowledge.

\begin{table}[h]
    \centering
    \begin{tabular}{l|c}
        \toprule
            \textbf{Model} &  \textbf{Dev} \\ \midrule 
            Transformer-XL & 19.0 \\ 
            Triple - Relation - Tail & 19.0 \\
            Triple - Relation & 18.7\\
            Triple & \textbf{18.5} \\ 
        \bottomrule
    \end{tabular}
    \caption{Ablating relation and/or tail entity from a relation triple. \label{tab:ablate_relation_triple}}
\end{table}

\begin{figure}[h]
    \centering
    \includegraphics[width=0.8\linewidth]{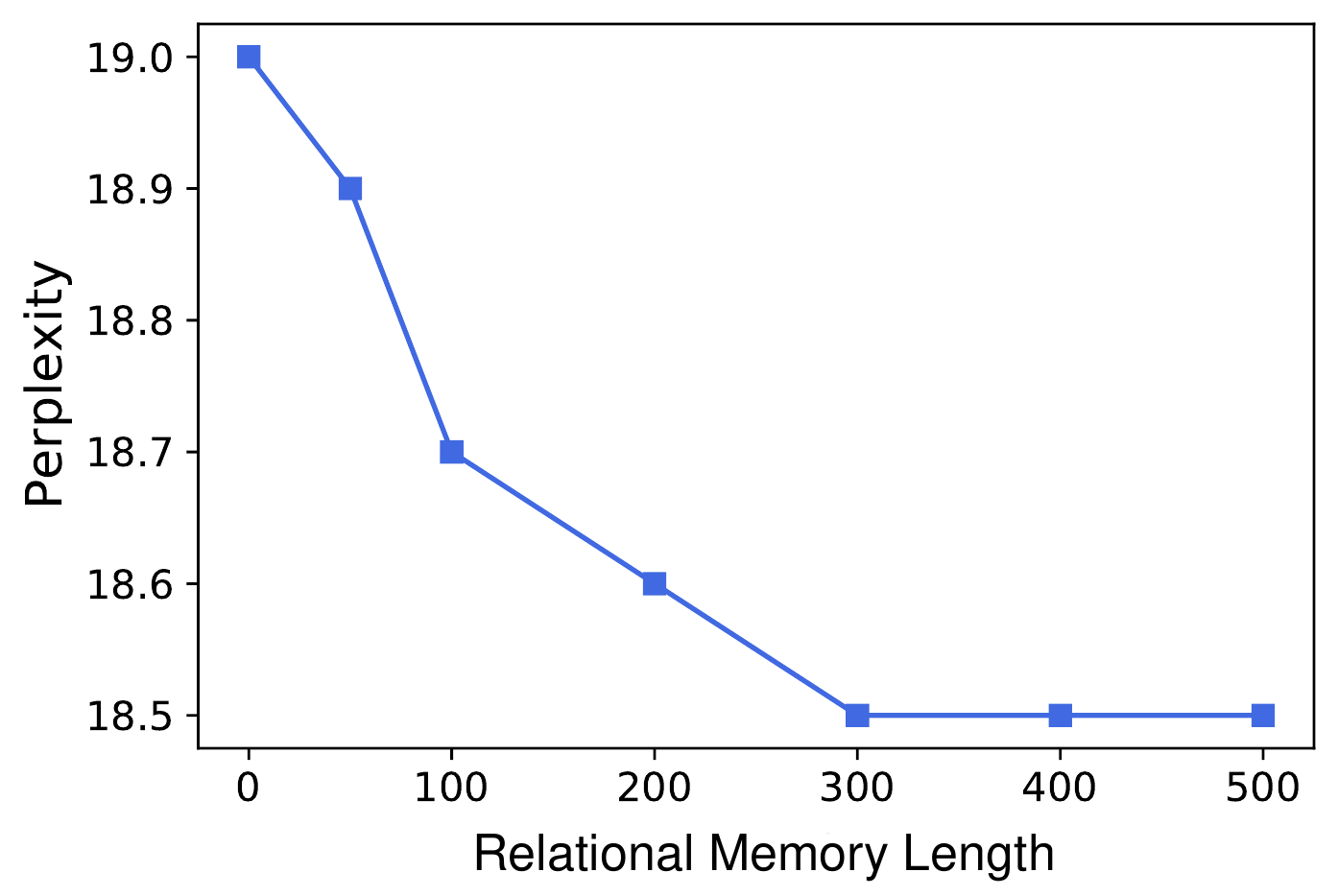}
    \vspace{-1em}
    \caption{Perplexity on WikiText-103 with different number of relation triples. }
    \label{fig:relation_perplexity}
\end{figure}

\paragraph{Length of relational memory.}
We study how many relation triples need to be stored in the relational memory. As shown in Figure \ref{fig:relation_perplexity}, we can see that the perplexity improves with more relation triples. However, the curve becomes flat with more than 300 relation triples. 

\begin{figure}[h]
    \centering
    \includegraphics[width=0.9\linewidth]{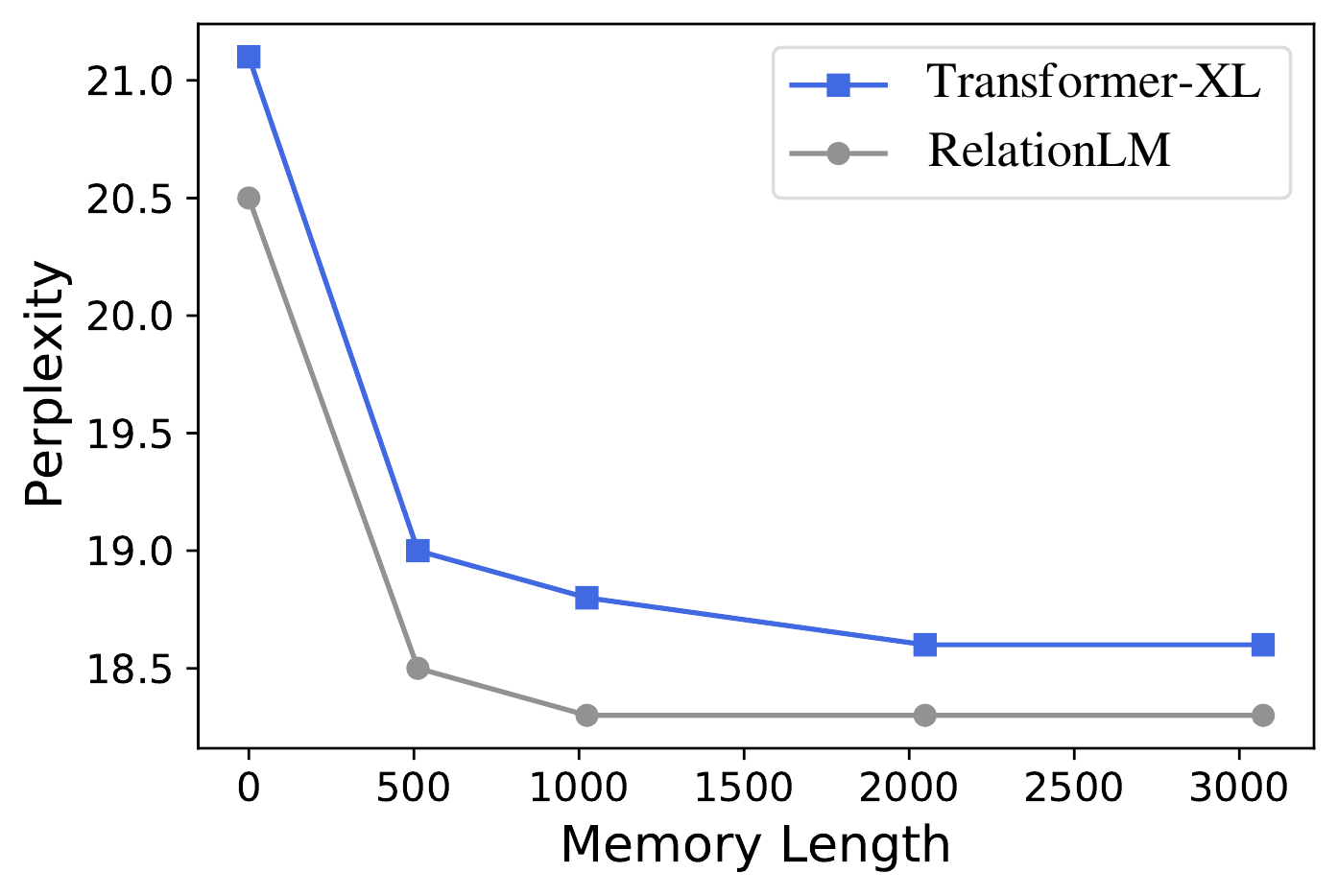}
    \vspace{-1.4em}
    \caption{Increasing extended memory length.}
    \label{fig:transformer_mem_len}
\end{figure}

\paragraph{Length of transformer-XL memory.}
As increasing the length of context window can capture longer dependency, we study whether increasing the length of extended (transformer-XL) memory removes the performance gap between \textsc{RelationLM} and transformer-XL. As shown in Figure \ref{fig:transformer_mem_len}, the performance of both \textsc{RelationLM} and transformer-XL improves with larger extended memory. However, \textsc{RelationLM} still outperforms transformer-XL even with extended memory length 3072. We conclude that relational memory brings complementary benefits to simply expanding extended memory, since it provides global information about entities on each dataset.

\paragraph{Dynamic OpenIE.}
All our main results use dynamic OpenIE. We show results without dynamic OpenIE in Table~\ref{tab:dynamic_openie}. We include the results on three datasets for a comparison. We can see that \textsc{RelationLM} with dynamic OpenIE performs comparably to \textsc{RelationLM} without dynamic OpenIE on WikiText-103 and enwik8, while larger improvements are obtained on WMT19. This indicates that dynamic OpenIE is more helpful for the news domain, which is more dynamic and temporal compared to knowledge-driven articles.


\begin{table}[h]
\centering
    \begin{tabular}{c|c|c|c}
        \toprule
            \textbf{Model} & \textbf{Wiki} & \textbf{WMT} & \textbf{ew8} \\ \midrule
             Transformer-XL & 19.0 & 21.7 & 1.05\\ 
            w/o Dynamic OpenIE &  18.6 & 21.4 & \textbf{1.04} \\  
            w/ Dynamic OpenIE & \textbf{18.5} & \textbf{21.0} & \textbf{1.04} \\
        \bottomrule
    \end{tabular}
    \caption{Perplexity with and without dynamic OpenIE. \label{tab:dynamic_openie}}

\end{table}

\paragraph{Entity scoring}

We study different entity scoring mechanisms for relation retrieval. We consider random selection (where entities extracted from the observed context are randomly selected), frequency-based scoring, and tf-idf scoring. As shown in Table \ref{tab:entity_scoring}, tf-idf performs the best.  

\begin{table}[h]
\centering
    \begin{tabular}{l|c}
        \toprule
            \textbf{Model} & \textbf{Dev}\\ \midrule
            Random & 19.1 \\ 
            Frequency & 18.7 \\ 
            tf-idf & \textbf{18.5}\\ 
        \bottomrule
    \end{tabular}
    \caption{Perplexity with different entity scoring methods.\label{tab:entity_scoring}}
\end{table}

\paragraph{Speed comparison.}
The wall clock time for both training and evaluation is shown in Table \ref{tab:time}. \textsc{RelationLM} is 1.5 and 2.1 times slower during training and evaluation, respectively. Evaluation slows down some more due to dynamic OpenIE as shown in Algorithm \ref{al:relation}.

\begin{table}[h]
    \centering
    \begin{tabular}{l|c|c}
        \toprule
            \textbf{Model} & \textbf{Train} & \textbf{Eval} \\ \midrule
            Transformer-XL & \textbf{0.51} & \textbf{0.31}\\
            \textsc{RelationLM} & 0.76 & 0.65\\
        \bottomrule
    \end{tabular}
    \caption{The unit is second/step. We use batch size 128 and 1 per step for training and evaluation, respectively. \label{tab:time}}
\end{table}

\begin{table}[t]
    \centering
    \small
    \begin{tabular}{l|c|c|c}
        \toprule
            \textbf{Dataset} & \textbf{Subset} & \textbf{\# Entity} & \textbf{\# Non-Entity} \\ \midrule
        \multirow{2}{*}{\small WikiText} & Dev & 61.6K & 155.9K \\
            & Test & 65.8K & 179.7K \\ \midrule
        \multirow{2}{*}{\small WMT} & Dev & 84.9K & 262.2K \\
        & Test &  81.0K & 256.6K\\ \midrule
        \multirow{2}{*}{\small enwik8} & Dev & 1.7M & 3.3M \\
        & Test & 1.7M & 3.3M \\
        \bottomrule
    \end{tabular}
    \caption{Statistics of entity and non-entity tokens. \label{tab:entiy_non_entity_stat}}
\end{table}
\begin{table}[t]
    \centering
    \small
    \begin{tabular}{l|l|l|c|c}
        \toprule
            & \textbf{Metric} & \textbf{Model} & \textbf{Dev} & \textbf{Test} \\ \midrule
            \multirow{6}{*}{\rotatebox[origin=c]{90}{\small Knowledge PPX}} & \multirow{2}{*}{WikiText} & Transformer-XL & 47.3 & 52.3\\
            & & \textsc{RelationLM} & \textbf{45.6} & \textbf{50.9}\\  \cmidrule{2-5}
            & \multirow{2}{*}{WMT} & Transformer-XL & 77.2 & 77.0\\
            & & \textsc{RelationLM} & \textbf{73.2} & \textbf{73.1}\\ \cmidrule{2-5}
            & \multirow{2}{*}{enwik8} & Transformer-XL & 2.25 & 2.21\\
            & & \textsc{RelationLM} & \textbf{2.22} & \textbf{2.19} \\ \cmidrule{2-5}
            \cmidrule{1-5}
            \multirow{6}{*}{\rotatebox[origin=c]{90}{Non-entity PPX}} & \multirow{2}{*}{WikiText} & Transformer-XL & 13.3 & 13.8\\
            & & \textsc{RelationLM} & \textbf{13.0} & \textbf{13.4}\\
            \cmidrule{2-5}
            & \multirow{2}{*}{WMT} & Transformer-XL & 14.4 & 14.4\\
            & & \textsc{RelationLM} & \textbf{14.2} & \textbf{14.3}\\ \cmidrule{2-5}
             
            & \multirow{2}{*}{enwik8} & Transformer-XL & \textbf{1.98} & \textbf{1.95}\\
            & & \textsc{RelationLM} & \textbf{1.98} & \textbf{1.95}\\   
            \bottomrule
    \end{tabular}
    \caption{Knowledge perplexity ($\downarrow$) and non-entity perplexity ($\downarrow$). \label{tab:hallucination}}
\end{table}

\begin{table}[t]
    \centering
    \begin{tabular}{l|l|c|c}
        \toprule
             \textbf{Metric} & \textbf{Model} & \textbf{Dev} & \textbf{Test} \\ \midrule
             \multirow{2}{*}{WikiText} & Transformer-XL &  9.9 & 9.4 \\
            & \textsc{RelationLM} & \textbf{11.4} & \textbf{11.2} \\ \cmidrule{1-4}
            \multirow{2}{*}{WMT} & Transformer-XL & 11.4 & 11.0\\
            & \textsc{RelationLM} & \textbf{12.6} & \textbf{12.3}\\ \cmidrule{1-4}
            \multirow{2}{*}{enwik8} & Transformer-XL & 16.0 & 18.9\\
            & \textsc{RelationLM} & \textbf{16.6} & \textbf{19.4}\\    
            \bottomrule
    \end{tabular}
    \caption{Knowledge $F_1$ ($\uparrow$).  \label{tab:knowledge_f1}}
\end{table}

\subsection{Does Relational Memory Improve Coherence? \label{sec:hallucination}}

For evaluating coherence, we use two automatic metrics---knowledge perplexity and knowledge $F_1$---to investigate whether the models can faithfully use entities. We further perform a human evaluation to study whether language models can generate coherent and knowledgeable sequences. We believe the human evaluation is a reliable way of evaluating coherence. This claim is advocated in \citet{DBLP:conf/acl/BarzilayL05}. We note that question answering is also often used to evaluate coherence \cite{DBLP:journals/corr/abs-2002-08909,DBLP:journals/corr/abs-2109-07958}. We leave this to future work.

\paragraph{Knowledge perplexity.} While vanilla perplexity considers all words in an evaluation set, knowledge perplexity only considers entities for calculating perplexity. We use it to evaluate whether the model can assign higher probabilities for the correct entities under different contexts. Table~\ref{tab:entiy_non_entity_stat} shows the numbers of entity words and non-entity words in our corpora. We show the results in Table~\ref{tab:hallucination}. We observe that the gap between \textsc{RelationLM} and transformer-XL is larger on knowledge perplexity. \textsc{RelationLM} only performs comparably or slightly better compared to transformer-XL on non-entity perplexity. This shows that relational memory is helpful for predicting entity words. Note that 
knowledge perplexity tends to be much higher 
than perplexity on non-entity words, indicating the difficulty of predicting entity words. 
This collection of results indicates that 
relational memory helps the model use entities coherently and consistently under different contexts.

\paragraph{Knowledge $F_1$.} We use knowledge $F_1$ to explore whether our model generates tokens that are grounded to its contexts. Given a context as input, we sequentially generate 32 words (or 128 characters) for word-(character-)level language modelling by sampling from the distribution of the next word (character). 
To reduce variance, we generate 100 continuations for each context. We then perform entity recognition for both the generated sequences and their corresponding ground-truth sequences and calculate an $F_1$ score based on these two sets of entities. For example, given the context ``...Ayola was nominated and shortlisted for the `Female Performance in TV' award'', we compare the generated text and the ground truth ``in the 2006 Screen Nation Awards, for her role as Kyla Tyson in Holby City...'' to calculate $F_1$. The results are shown in Table \ref{tab:knowledge_f1}. We notice that \textsc{RelationLM} performs better compared to transformer-XL. We conclude that models with relational memory can generate more coherent and logical text. 


\paragraph{Human evaluation.} 

We conduct a human evaluation to study whether language models can generate coherent and knowledgeable sequences. We take 1,000 contexts from the test set of WikiText-103. We show the contexts, ground-truth sequences, and continuations generated by \textsc{RelationLM} and transformer-XL to five annotators. We use greedy decoding for both models. We shuffle the order of the continuations generated by \textsc{RelationLM} and transformer-XL so that the annotators are unaware of the sources of sequences. We then pose the following questions to the annotators:

\begin{enumerate}
    \item \textbf{Coherent.} Given the context and its ground-truth continuation for reference, which generated sequence is more logical and coherent?
    
    \item \textbf{Knowledgeable.} Given the context and its ground-truth continuation, which generated sequence provides more insights and is more knowledgeable?
\end{enumerate}

We show the results in Table \ref{tab:human_eval}. We find that \textsc{RelationLM} outperforms transformer-XL in the human evaluation. These results are consistent with the two automatic metrics, knowledge perplexity and knowledge $F_1$. This corroborates our claim that relational memory improves coherence in language modelling.

\begin{table}[h]
    \centering
    \begin{tabular}{l|c|c}
        \toprule
             \textbf{Model} & \textbf{Coherent} & \textbf{Knowledgeable} \\ \midrule
             Transformer-XL & 388 & 416\\
             \textsc{RelationLM} & \textbf{612} & \textbf{584}\\
            \bottomrule
    \end{tabular}
    \caption{We show the number of contexts in which a continuation from a particular model is chosen by human evaluators for each evaluation criterion. Recall that the total number of contexts used for human evaluation is 1,000. Since we have five annotators, we use majority voting to decide the favored model for each continuation. We use the Kappa statistic to measure inter-annotator agreement. The statistic is 0.64, which shows substantial agreement among the annotators.  \label{tab:human_eval}}
\end{table}


\subsection{Qualitative Analysis}
\label{sec:qualitative}

\begin{figure*}[t]
    \centering
    \includegraphics[width=1\textwidth, height=90pt]{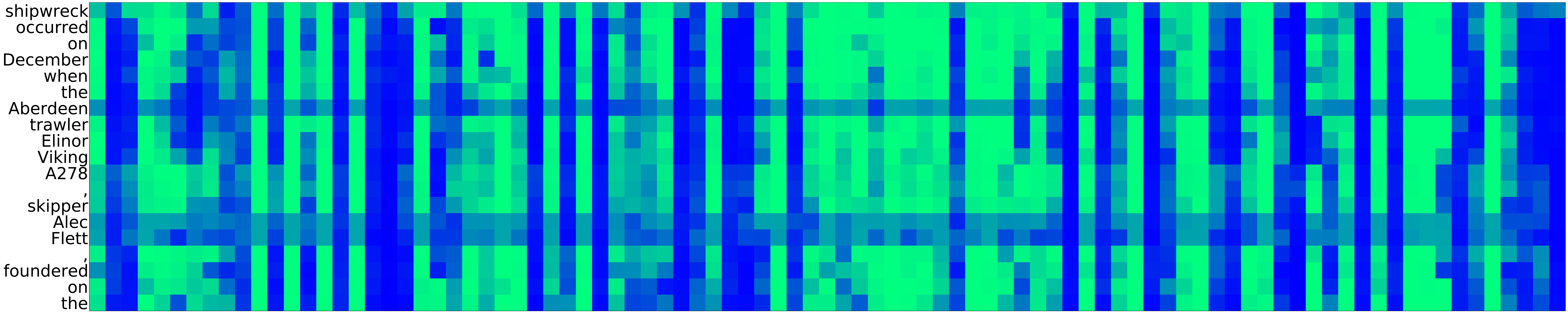}
    \caption{Heatmap of gate values. }
    \label{fig:heatmap}
\end{figure*}
\begin{figure}[h]
    \centering
    \includegraphics[width=0.8\linewidth]{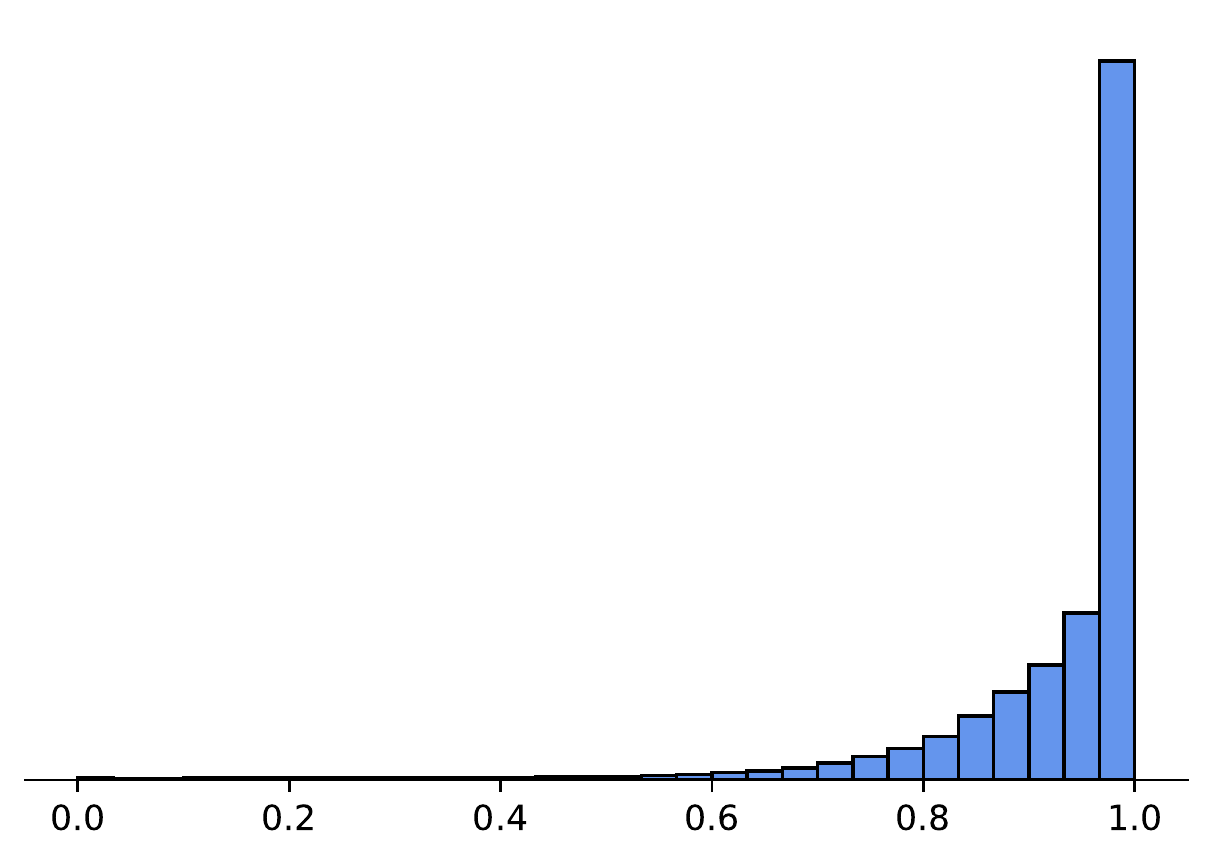}
    \caption{Histogram of gate values $\mathbf{g}_t$.}
    \label{fig:gate_distribution}
\end{figure}

\paragraph{Gate values.}
As we use a gating function to integrate transformer-XL with relational information, we study gate values in this section. The histogram of gate values is shown in Figure \ref{fig:gate_distribution}. We notice that the histogram concentrates around 0.9. This is expected because non-entity words, which account for a large portion of text (according to Table \ref{tab:entiy_non_entity_stat}), benefit less from the relational memory and mainly rely on the observed context for prediction as shown in \S\ref{sec:hallucination}. We further calculate the average gate values for entity words and non-entity words. The average gate value for entity words is 0.87, while the average value is 0.92 for non-entity words. This confirms that entity words rely more on relational information for prediction compared to non-entity words. We also plot the heatmap of gate values and a cherry-picked example is shown in Figure \ref{fig:heatmap}. Note that we randomly select 100 dimensions from 512 dimensions for readability. We notice that the entities, Aberdeen and Alec Flett, use more relational information than other positions (as shown by the horizontal blue lines). These results demonstrate that \textsc{RelationLM} can adaptively incorporate relational information for prediction. 


\paragraph{Example.} We show three cherry-picked examples in Table~\ref{fig:case_study}. We take the first for illustration, which shows a text segment from the article, Joe Biden 2008 presidential campaign\footnote{\url{https://en.wikipedia.org/wiki/Joe_Biden_2008_presidential_campaign}} and some retrieved relations. We find that the first two relations, (Joe Biden, senior Senator, Delaware) and (Joe Biden presidential campaign, began, January 7 2007), are extracted from previous text segments, while (Joe Biden, was nominated, vice president) and (Biden, withdrew nomination, 1987) are extracted from the other articles, Joe Biden\footnote{\url{https://en.wikipedia.org/wiki/Joe_Biden}} and Joe Biden 1988 presidential campaign\footnote{\url{https://en.wikipedia.org/wiki/Joe_Biden_1988_presidential_campaign}}, respectively. We notice that the relation (Joe Biden, was nominated, vice president) is highly predictive of the sequence, ``Biden was selected to
be Democratic presidential nominee Barack Obama's vice presidential running mate''. From the observed context, the model also identifies a closely related entity, Barack Obama, and retrieves the relation (Barack Obama, president of, United States). Therefore, we conclude that the relational memory can give a global picture of related entities and provide relevant information for language modelling. 

\paragraph{Causal intervention.} We use causal intervention to study whether changing the contents in the relational memory will affect language model prediction. Given the relation (Obama, born in, Hawaii) along with other relations about Barack Obama, we let the model complete the sequence, ``Obama was born in''. \textsc{RelationLM} outputs ``Obama was born in and raised in Hawaii.'' with greedy decoding. However, after modifying the relation to (Obama, born in, Kenya), we obtain ``Obama was born in Kenya and was the first African-American president.''. We further change to (Obama, born in, Paris) and the model outputs ``Obama was born in Paris, France.''. This indicates that \textsc{RelationLM} can take advantage of relation triples for making prediction. While we can also use prompts as intervention for vanilla language models, it remains challenging about selecting the appropriate prompts in different applications \cite{DBLP:journals/corr/abs-2107-13586}.

\begin{table}[t]
\setlength{\arrayrulewidth}{.1em}
\begin{tabular}{c}
    \toprule
     \includegraphics[width=\linewidth]{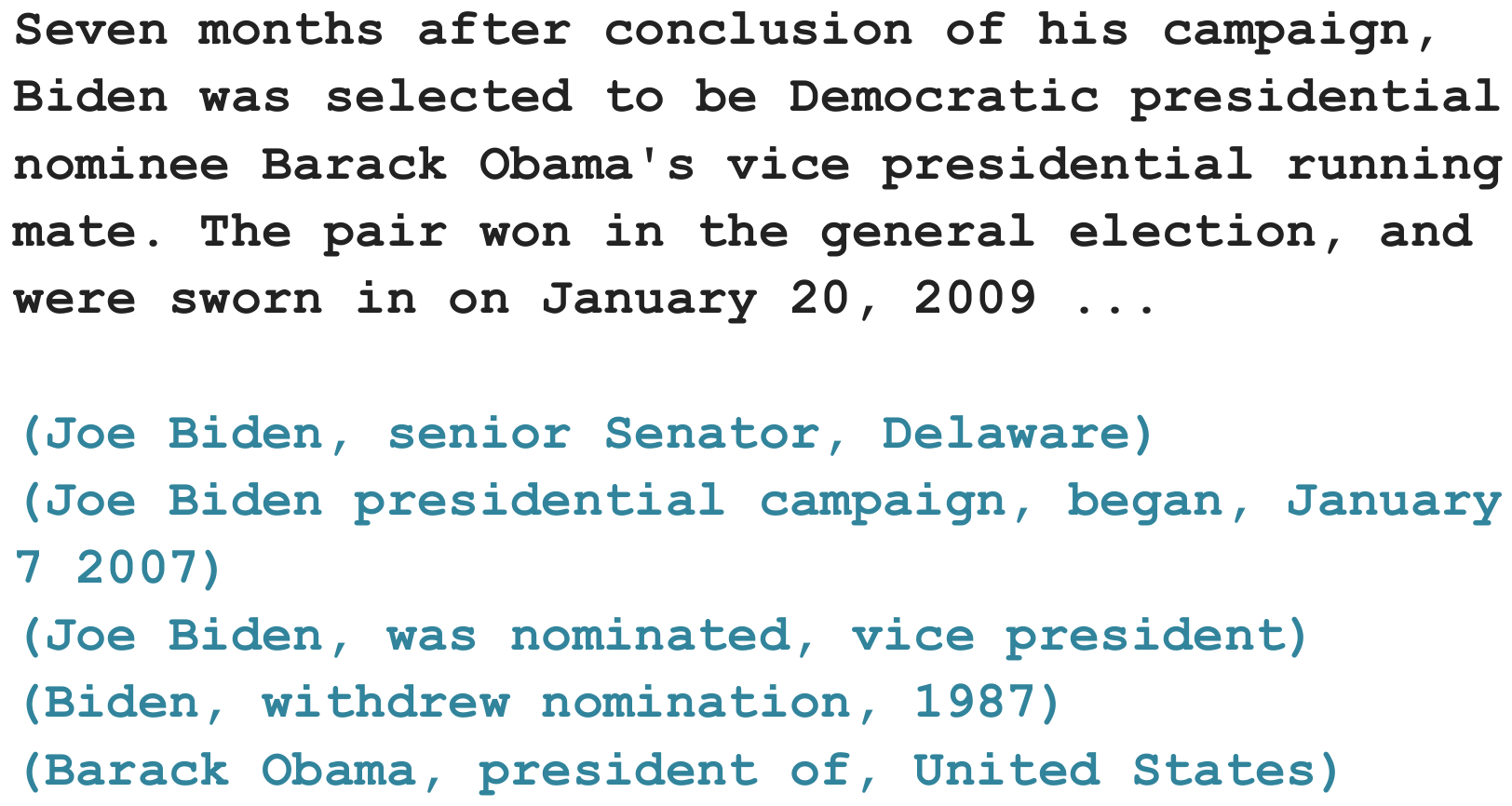} \\
     \midrule
     \includegraphics[width=\linewidth]{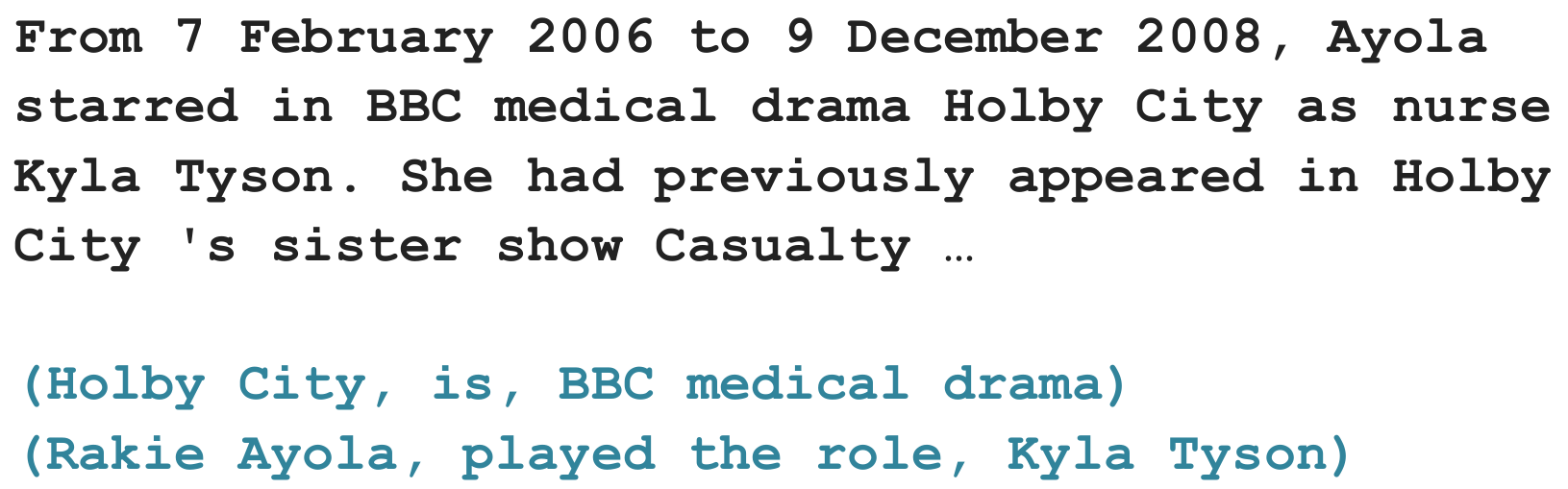} \\
     \midrule
     \includegraphics[width=\linewidth]{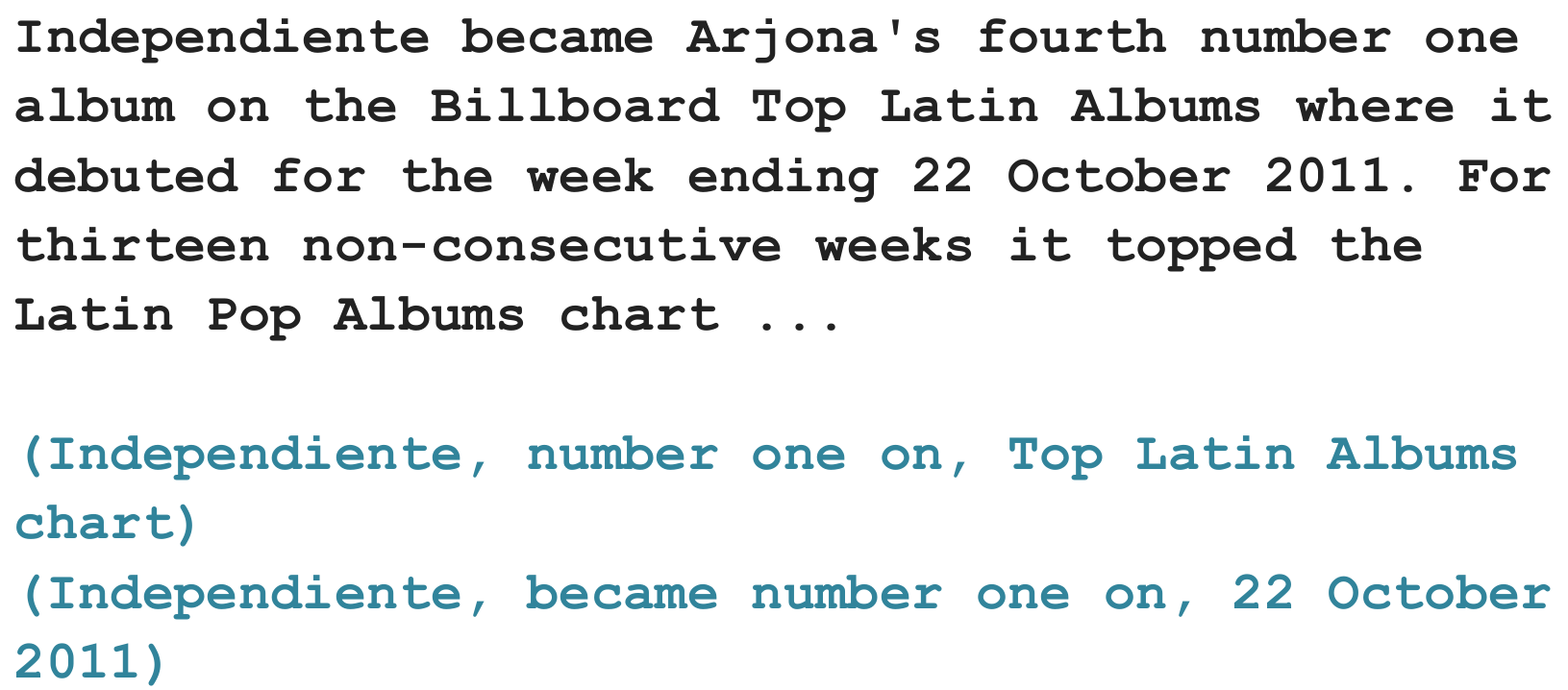} \\
     \bottomrule
\end{tabular}
\caption{Three examples of text segment and retrieved relations (based on previous text segments).\label{fig:case_study}}
\end{table}

\section{Related Work}

\paragraph{Knowledge-enhanced architectures.} Injecting symbolic knowledge to machine learning models is widely-adopted to improve the performance of natural language understanding \cite{DBLP:conf/naacl/AnnervazCD18,DBLP:conf/konvens/OstendorffBBSRG19}, question answering \cite{DBLP:conf/aaai/ZhangDKSS18,DBLP:conf/wsdm/HuangZLL19,DBLP:conf/naacl/HixonCH15}, dialogue systems \cite{DBLP:conf/acl/WuLCZDYZL18,Moon2019opendialkg,DBLP:conf/nips/GuoTDZY18,liu-etal-2021-pretraining} and recommendation systems \cite{DBLP:conf/kdd/ZhangYLXM16,DBLP:conf/www/WangZXG18,DBLP:conf/www/WangZZLXG19}. Different from these models, we focus on using symbolic knowledge for language modelling. Existing language models are prone to generating illogical and contradictory contents. We believe that connecting language modelling and knowledge graphs is a promising direction to overcome the problem. Next we review previous knowledge-enhanced language models.


\paragraph{Knowledge-enhanced language models.} Our model is closely related to previous work on grounding autoregressive language models with knowledge graphs \cite{ahn2016neural,DBLP:conf/acl/LoganLPGS19,DBLP:conf/aaai/HayashiHXN20,wang-etal-2021-wikigraphs}. However, these models rely on complex and adhoc preprocessing or rules to link text with knowledge bases, e.g.\ Freebase and Wikidata. As a result, previous work is more aligned with conditional language modelling, e.g.\ graph-to-text generation $ p(\mathbf{x} | \mathcal{G}) $ in \citet{wang-etal-2021-wikigraphs}, which contrasts with unconditional language modeling $ p(\mathbf{x}) $ considered in this work. As the graph $\mathcal{G}$ is constructed with the \textit{unseen} text $\mathbf{x}$, predicting $\mathbf{x}$ given $\mathcal{G}$ is easier due to this information leakage for \citet{wang-etal-2021-wikigraphs}. Also in \citet{DBLP:conf/aaai/HayashiHXN20}, topic entities are required for language modelling, which may not be available in most datasets, e.g.\ the news domain. We do not compare with these previous models due to the different settings. In contrast, we adopt OpenIE relations and use a tf-idf search to retrieve relation triples for connecting language models and knowledge graphs. In the experiments, we demonstrate the effectiveness of our approach on three datasets, WikiText-103, WMT19 and enwik8.

There are language models incorporating entity information, such as entity coreference annotations \cite{DBLP:conf/emnlp/JiTMCS17,DBLP:conf/naacl/ClarkJS18}, surface forms of entities \cite{DBLP:conf/emnlp/KiddonZC16,DBLP:conf/emnlp/YangBDL17,DBLP:conf/iclr/CaoI0P21}, entity types \cite{DBLP:conf/acl/ChangPCR18,DBLP:conf/inlg/0005PHZJJK18} and entity descriptions \cite{DBLP:journals/corr/BahdanauBJGVB17}. Different from these models, we augment language models with a relational memory consisting of relation triples. We demonstrate the effectiveness of using relation triples by ablating tail entities and relations in \S\ref{sec:ablation}.

\paragraph{Knowledge-enhanced pretraining.} Using knowledge information for pretraining language models \cite{DBLP:conf/emnlp/PetersNLSJSS19,DBLP:journals/corr/abs-1904-09223,DBLP:conf/aaai/LiuZ0WJD020,DBLP:journals/corr/abs-2002-08909,wang2021kepler,agarwal2021knowledge,DBLP:conf/naacl/VergaSSC21} has recently grown in popularity and has achieved substantial improvements on knowledge-driven tasks such as question answering and named entity recognition. Instead of using knowledge information for improving downstream knowledge-driven tasks, we focus on using knowledge information for improving the generation capability of the language model itself.

\paragraph{Retrieval-augmented models.}
Retrieval-augmented models are now widely adopted in open-domain question answering \cite{DBLP:conf/acl/ChenFWB17,DBLP:conf/nips/LewisPPPKGKLYR020,mbpa2019,DBLP:conf/eacl/IzacardG21}, dialogue \cite{DBLP:conf/iclr/DinanRSFAW19,DBLP:journals/tacl/FanGBB21,thulke2021efficient} and machine translation \cite{DBLP:conf/naacl/BapnaF19,DBLP:journals/corr/abs-2010-00710}. We focus on retrieval augmentation for language modelling \cite{DBLP:conf/iclr/MerityX0S17,DBLP:journals/corr/GraveJU16,DBLP:conf/iclr/KhandelwalLJZL20,DBLP:journals/tacl/YogatamadK21}. These algorithms are specifically tailored for language modelling, where related tokens are retrieved to help predict the next token. In this work, we move beyond token augmentation and show the benefits of retrieving relation triples. We also demonstrate that our model is complementary to a token augmentation model, \textsc{Spalm} \cite{DBLP:journals/tacl/YogatamadK21} in the experiments.

\section{Conclusion}

We presented \textsc{RelationLM}, a language model that is augmented with relational memory.
We showed how to obtain relevant knowledge graphs for a given corpus
and how to combine them with a state-of-the-art language model such as transformer-XL.
We demonstrated that our model improves performance and coherence on WikiText-103, WMT19 and enwik8.
We also performed a comprehensive analysis to better understand how our model works.
Our model provides a way to combine an autoregressive language model with general knowledge graphs.

\section*{Acknowledgements}
We would like to thank our action editor (Xavier Carreras) and three anonymous reviewers for their insightful comments. We also thank Angeliki Lazaridou, Cyprien de Masson d'Autume, Lingpeng Kong, Laura Rimell, Aida Nematzadeh, and the DeepMind language team for their helpful discussions.

\bibliography{tacl2018}

\begin{thebibliography}{90}
\expandafter\ifx\csname natexlab\endcsname\relax\def\natexlab#1{#1}\fi

\bibitem[{Agarwal et~al.(2021)Agarwal, Ge, Shakeri, and
  Al-Rfou}]{agarwal2021knowledge}
Oshin Agarwal, Heming Ge, Siamak Shakeri, and Rami Al-Rfou. 2021.
\newblock Knowledge graph based synthetic corpus generation for
  knowledge-enhanced language model pre-training.
\newblock In \emph{Proceedings of the 2021 Conference of the North American
  Chapter of the Association for Computational Linguistics: Human Language
  Technologies}, pages 3554--3565.

\bibitem[{Ahn et~al.(2016)Ahn, Choi, P{\"a}rnamaa, and Bengio}]{ahn2016neural}
Sungjin Ahn, Heeyoul Choi, Tanel P{\"a}rnamaa, and Yoshua Bengio. 2016.
\newblock A neural knowledge language model.
\newblock \emph{arXiv preprint arXiv:1608.00318}.

\bibitem[{Angeli et~al.(2015)Angeli, Premkumar, and
  Manning}]{DBLP:conf/acl/AngeliPM15}
Gabor Angeli, Melvin Jose~Johnson Premkumar, and Christopher~D. Manning. 2015.
\newblock \href {https://doi.org/10.3115/v1/p15-1034} {Leveraging linguistic
  structure for open domain information extraction}.
\newblock In \emph{Proceedings of the 53rd Annual Meeting of the Association
  for Computational Linguistics and the 7th International Joint Conference on
  Natural Language Processing of the Asian Federation of Natural Language
  Processing, {ACL} 2015, July 26-31, 2015, Beijing, China, Volume 1: Long
  Papers}, pages 344--354. The Association for Computer Linguistics.

\bibitem[{Annervaz et~al.(2018)Annervaz, Chowdhury, and
  Dukkipati}]{DBLP:conf/naacl/AnnervazCD18}
K.~M. Annervaz, Somnath Basu~Roy Chowdhury, and Ambedkar Dukkipati. 2018.
\newblock \href {https://doi.org/10.18653/v1/n18-1029} {Learning beyond
  datasets: Knowledge graph augmented neural networks for natural language
  processing}.
\newblock In \emph{Proceedings of the 2018 Conference of the North American
  Chapter of the Association for Computational Linguistics: Human Language
  Technologies, {NAACL-HLT} 2018, New Orleans, Louisiana, USA, June 1-6, 2018,
  Volume 1 (Long Papers)}, pages 313--322. Association for Computational
  Linguistics.

\bibitem[{Bahdanau et~al.(2017)Bahdanau, Bosc, Jastrzebski, Grefenstette,
  Vincent, and Bengio}]{DBLP:journals/corr/BahdanauBJGVB17}
Dzmitry Bahdanau, Tom Bosc, Stanislaw Jastrzebski, Edward Grefenstette, Pascal
  Vincent, and Yoshua Bengio. 2017.
\newblock \href {http://arxiv.org/abs/1706.00286} {Learning to compute word
  embeddings on the fly}.
\newblock \emph{CoRR}, abs/1706.00286.

\bibitem[{Bapna and Firat(2019)}]{DBLP:conf/naacl/BapnaF19}
Ankur Bapna and Orhan Firat. 2019.
\newblock \href {https://doi.org/10.18653/v1/n19-1191} {Non-parametric
  adaptation for neural machine translation}.
\newblock In \emph{Proceedings of the 2019 Conference of the North American
  Chapter of the Association for Computational Linguistics: Human Language
  Technologies, {NAACL-HLT} 2019, Minneapolis, MN, USA, June 2-7, 2019, Volume
  1 (Long and Short Papers)}, pages 1921--1931. Association for Computational
  Linguistics.

\bibitem[{Barrault et~al.(2019)Barrault, Bojar, Costa-juss{\`a}, Federmann,
  Fishel, Graham, Haddow, Huck, Koehn, Malmasi, Monz, M{\"u}ller, Pal, Post,
  and Zampieri}]{barrault-etal-2019-findings}
Lo{\"\i}c Barrault, Ond{\v{r}}ej Bojar, Marta~R. Costa-juss{\`a}, Christian
  Federmann, Mark Fishel, Yvette Graham, Barry Haddow, Matthias Huck, Philipp
  Koehn, Shervin Malmasi, Christof Monz, Mathias M{\"u}ller, Santanu Pal, Matt
  Post, and Marcos Zampieri. 2019.
\newblock \href {https://doi.org/10.18653/v1/W19-5301} {Findings of the 2019
  conference on machine translation ({WMT}19)}.
\newblock In \emph{Proceedings of the Fourth Conference on Machine Translation
  (Volume 2: Shared Task Papers, Day 1)}, pages 1--61, Florence, Italy.
  Association for Computational Linguistics.

\bibitem[{Barzilay and Lapata(2005)}]{DBLP:conf/acl/BarzilayL05}
Regina Barzilay and Mirella Lapata. 2005.
\newblock \href {https://doi.org/10.3115/1219840.1219858} {Modeling local
  coherence: An entity-based approach}.
\newblock In \emph{{ACL} 2005, 43rd Annual Meeting of the Association for
  Computational Linguistics, Proceedings of the Conference, 25-30 June 2005,
  University of Michigan, {USA}}, pages 141--148. The Association for Computer
  Linguistics.

\bibitem[{Bengio et~al.(2003)Bengio, Ducharme, Vincent, and
  Janvin}]{DBLP:journals/jmlr/BengioDVJ03}
Yoshua Bengio, R{\'{e}}jean Ducharme, Pascal Vincent, and Christian Janvin.
  2003.
\newblock \href {http://jmlr.org/papers/v3/bengio03a.html} {A neural
  probabilistic language model}.
\newblock \emph{J. Mach. Learn. Res.}, 3:1137--1155.

\bibitem[{Bollacker et~al.(2007)Bollacker, Cook, and
  Tufts}]{DBLP:conf/aaai/BollackerCT07}
Kurt~D. Bollacker, Robert~P. Cook, and Patrick Tufts. 2007.
\newblock \href {http://www.aaai.org/Library/AAAI/2007/aaai07-355.php}
  {Freebase: {A} shared database of structured general human knowledge}.
\newblock In \emph{Proceedings of the Twenty-Second {AAAI} Conference on
  Artificial Intelligence, July 22-26, 2007, Vancouver, British Columbia,
  Canada}, pages 1962--1963. {AAAI} Press.

\bibitem[{Bordes et~al.(2013)Bordes, Usunier, Garc{\'{\i}}a{-}Dur{\'{a}}n,
  Weston, and Yakhnenko}]{DBLP:conf/nips/BordesUGWY13}
Antoine Bordes, Nicolas Usunier, Alberto Garc{\'{\i}}a{-}Dur{\'{a}}n, Jason
  Weston, and Oksana Yakhnenko. 2013.
\newblock \href
  {https://proceedings.neurips.cc/paper/2013/hash/1cecc7a77928ca8133fa24680a88d2f9-Abstract.html}
  {Translating embeddings for modeling multi-relational data}.
\newblock In \emph{Advances in Neural Information Processing Systems 26: 27th
  Annual Conference on Neural Information Processing Systems 2013. Proceedings
  of a meeting held December 5-8, 2013, Lake Tahoe, Nevada, United States},
  pages 2787--2795.

\bibitem[{Bradbury et~al.(2018)Bradbury, Frostig, Hawkins, Johnson, Leary,
  Maclaurin, and Wanderman-Milne}]{jax2018github}
James Bradbury, Roy Frostig, Peter Hawkins, Matthew~James Johnson, Chris Leary,
  Dougal Maclaurin, and Skye Wanderman-Milne. 2018.
\newblock \href {http://github.com/google/jax} {{JAX}: composable
  transformations of {P}ython+{N}um{P}y programs}.

\bibitem[{Brown et~al.(2020)Brown, Mann, Ryder, Subbiah, Kaplan, Dhariwal,
  Neelakantan, Shyam, Sastry, Askell et~al.}]{brown2020language}
Tom~B Brown, Benjamin Mann, Nick Ryder, Melanie Subbiah, Jared Kaplan, Prafulla
  Dhariwal, Arvind Neelakantan, Pranav Shyam, Girish Sastry, Amanda Askell,
  et~al. 2020.
\newblock Language models are few-shot learners.
\newblock \emph{arXiv preprint arXiv:2005.14165}.

\bibitem[{Cao et~al.(2021)Cao, Izacard, Riedel, and
  Petroni}]{DBLP:conf/iclr/CaoI0P21}
Nicola~De Cao, Gautier Izacard, Sebastian Riedel, and Fabio Petroni. 2021.
\newblock \href {https://openreview.net/forum?id=5k8F6UU39V} {Autoregressive
  entity retrieval}.
\newblock In \emph{9th International Conference on Learning Representations,
  {ICLR} 2021, Virtual Event, Austria, May 3-7, 2021}. OpenReview.net.

\bibitem[{Chen et~al.(2017)Chen, Fisch, Weston, and
  Bordes}]{DBLP:conf/acl/ChenFWB17}
Danqi Chen, Adam Fisch, Jason Weston, and Antoine Bordes. 2017.
\newblock \href {https://doi.org/10.18653/v1/P17-1171} {Reading wikipedia to
  answer open-domain questions}.
\newblock In \emph{Proceedings of the 55th Annual Meeting of the Association
  for Computational Linguistics, {ACL} 2017, Vancouver, Canada, July 30 -
  August 4, Volume 1: Long Papers}, pages 1870--1879. Association for
  Computational Linguistics.

\bibitem[{Cho et~al.(2014)Cho, van Merrienboer, Bahdanau, and
  Bengio}]{DBLP:journals/corr/ChoMBB14}
KyungHyun Cho, Bart van Merrienboer, Dzmitry Bahdanau, and Yoshua Bengio. 2014.
\newblock \href {http://arxiv.org/abs/1409.1259} {On the properties of neural
  machine translation: Encoder-decoder approaches}.
\newblock \emph{CoRR}, abs/1409.1259.

\bibitem[{Clark et~al.(2018)Clark, Ji, and Smith}]{DBLP:conf/naacl/ClarkJS18}
Elizabeth Clark, Yangfeng Ji, and Noah~A. Smith. 2018.
\newblock \href {https://doi.org/10.18653/v1/n18-1204} {Neural text generation
  in stories using entity representations as context}.
\newblock In \emph{Proceedings of the 2018 Conference of the North American
  Chapter of the Association for Computational Linguistics: Human Language
  Technologies, {NAACL-HLT} 2018, New Orleans, Louisiana, USA, June 1-6, 2018,
  Volume 1 (Long Papers)}, pages 2250--2260. Association for Computational
  Linguistics.

\bibitem[{Dai et~al.(2019)Dai, Yang, Yang, Carbonell, Le, and
  Salakhutdinov}]{DBLP:conf/acl/DaiYYCLS19}
Zihang Dai, Zhilin Yang, Yiming Yang, Jaime~G. Carbonell, Quoc~Viet Le, and
  Ruslan Salakhutdinov. 2019.
\newblock \href {https://doi.org/10.18653/v1/p19-1285} {Transformer-xl:
  Attentive language models beyond a fixed-length context}.
\newblock In \emph{Proceedings of the 57th Conference of the Association for
  Computational Linguistics, {ACL} 2019, Florence, Italy, July 28- August 2,
  2019, Volume 1: Long Papers}, pages 2978--2988. Association for Computational
  Linguistics.

\bibitem[{Devlin et~al.(2019)Devlin, Chang, Lee, and
  Toutanova}]{DBLP:conf/naacl/DevlinCLT19}
Jacob Devlin, Ming{-}Wei Chang, Kenton Lee, and Kristina Toutanova. 2019.
\newblock \href {https://doi.org/10.18653/v1/n19-1423} {{BERT:} pre-training of
  deep bidirectional transformers for language understanding}.
\newblock In \emph{Proceedings of the 2019 Conference of the North American
  Chapter of the Association for Computational Linguistics: Human Language
  Technologies, {NAACL-HLT} 2019, Minneapolis, MN, USA, June 2-7, 2019, Volume
  1 (Long and Short Papers)}, pages 4171--4186. Association for Computational
  Linguistics.

\bibitem[{Dinan et~al.(2019)Dinan, Roller, Shuster, Fan, Auli, and
  Weston}]{DBLP:conf/iclr/DinanRSFAW19}
Emily Dinan, Stephen Roller, Kurt Shuster, Angela Fan, Michael Auli, and Jason
  Weston. 2019.
\newblock \href {https://openreview.net/forum?id=r1l73iRqKm} {Wizard of
  wikipedia: Knowledge-powered conversational agents}.
\newblock In \emph{7th International Conference on Learning Representations,
  {ICLR} 2019, New Orleans, LA, USA, May 6-9, 2019}. OpenReview.net.

\bibitem[{Etzioni et~al.(2008)Etzioni, Banko, Soderland, and
  Weld}]{etzioni2008open}
Oren Etzioni, Michele Banko, Stephen Soderland, and Daniel~S Weld. 2008.
\newblock Open information extraction from the web.
\newblock \emph{Communications of the ACM}, 51(12):68--74.

\bibitem[{Fan et~al.(2021)Fan, Gardent, Braud, and
  Bordes}]{DBLP:journals/tacl/FanGBB21}
Angela Fan, Claire Gardent, Chlo{\'{e}} Braud, and Antoine Bordes. 2021.
\newblock \href {https://transacl.org/ojs/index.php/tacl/article/view/2419}
  {Augmenting transformers with knn-based composite memory for dialog}.
\newblock \emph{Trans. Assoc. Comput. Linguistics}, 9:82--99.

\bibitem[{Grave et~al.(2017)Grave, Joulin, Ciss{\'e}, Grangier, and
  J{\'e}gou}]{pmlr-v70-grave17a}
{\'E}douard Grave, Armand Joulin, Moustapha Ciss{\'e}, David Grangier, and
  Herv{\'e} J{\'e}gou. 2017.
\newblock \href {http://proceedings.mlr.press/v70/grave17a.html} {Efficient
  softmax approximation for {GPU}s}.
\newblock In \emph{Proceedings of the 34th International Conference on Machine
  Learning}, volume~70 of \emph{Proceedings of Machine Learning Research},
  pages 1302--1310. PMLR.

\bibitem[{Grave et~al.(2016)Grave, Joulin, and
  Usunier}]{DBLP:journals/corr/GraveJU16}
Edouard Grave, Armand Joulin, and Nicolas Usunier. 2016.
\newblock \href {http://arxiv.org/abs/1612.04426} {Improving neural language
  models with a continuous cache}.
\newblock \emph{CoRR}, abs/1612.04426.

\bibitem[{Guo et~al.(2018)Guo, Tang, Duan, Zhou, and
  Yin}]{DBLP:conf/nips/GuoTDZY18}
Daya Guo, Duyu Tang, Nan Duan, Ming Zhou, and Jian Yin. 2018.
\newblock \href
  {https://proceedings.neurips.cc/paper/2018/hash/d63fbf8c3173730f82b150c5ef38b8ff-Abstract.html}
  {Dialog-to-action: Conversational question answering over a large-scale
  knowledge base}.
\newblock In \emph{Advances in Neural Information Processing Systems 31: Annual
  Conference on Neural Information Processing Systems 2018, NeurIPS 2018,
  December 3-8, 2018, Montr{\'{e}}al, Canada}, pages 2946--2955.

\bibitem[{Guu et~al.(2020)Guu, Lee, Tung, Pasupat, and
  Chang}]{DBLP:journals/corr/abs-2002-08909}
Kelvin Guu, Kenton Lee, Zora Tung, Panupong Pasupat, and Ming{-}Wei Chang.
  2020.
\newblock \href {http://arxiv.org/abs/2002.08909} {{REALM:} retrieval-augmented
  language model pre-training}.
\newblock \emph{CoRR}, abs/2002.08909.

\bibitem[{Hayashi et~al.(2020)Hayashi, Hu, Xiong, and
  Neubig}]{DBLP:conf/aaai/HayashiHXN20}
Hiroaki Hayashi, Zecong Hu, Chenyan Xiong, and Graham Neubig. 2020.
\newblock \href {https://aaai.org/ojs/index.php/AAAI/article/view/6298} {Latent
  relation language models}.
\newblock In \emph{The Thirty-Fourth {AAAI} Conference on Artificial
  Intelligence, {AAAI} 2020, The Thirty-Second Innovative Applications of
  Artificial Intelligence Conference, {IAAI} 2020, The Tenth {AAAI} Symposium
  on Educational Advances in Artificial Intelligence, {EAAI} 2020, New York,
  NY, USA, February 7-12, 2020}, pages 7911--7918. {AAAI} Press.

\bibitem[{Hendrycks and Gimpel(2016)}]{hendrycks2016gaussian}
Dan Hendrycks and Kevin Gimpel. 2016.
\newblock Gaussian error linear units (gelus).
\newblock \emph{arXiv preprint arXiv:1606.08415}.

\bibitem[{Hennigan et~al.(2020)Hennigan, Cai, Norman, and
  Babuschkin}]{haiku2020github}
Tom Hennigan, Trevor Cai, Tamara Norman, and Igor Babuschkin. 2020.
\newblock \href {http://github.com/deepmind/dm-haiku} {{H}aiku: {S}onnet for
  {JAX}}.

\bibitem[{Hixon et~al.(2015)Hixon, Clark, and
  Hajishirzi}]{DBLP:conf/naacl/HixonCH15}
Ben Hixon, Peter Clark, and Hannaneh Hajishirzi. 2015.
\newblock \href {https://doi.org/10.3115/v1/n15-1086} {Learning knowledge
  graphs for question answering through conversational dialog}.
\newblock In \emph{{NAACL} {HLT} 2015, The 2015 Conference of the North
  American Chapter of the Association for Computational Linguistics: Human
  Language Technologies, Denver, Colorado, USA, May 31 - June 5, 2015}, pages
  851--861. The Association for Computational Linguistics.

\bibitem[{Hochreiter and Schmidhuber(1997)}]{hochreiter1997long}
Sepp Hochreiter and J{\"u}rgen Schmidhuber. 1997.
\newblock Long short-term memory.
\newblock \emph{Neural computation}, 9(8):1735--1780.

\bibitem[{Huang et~al.(2019)Huang, Zhang, Li, and
  Li}]{DBLP:conf/wsdm/HuangZLL19}
Xiao Huang, Jingyuan Zhang, Dingcheng Li, and Ping Li. 2019.
\newblock \href {https://doi.org/10.1145/3289600.3290956} {Knowledge graph
  embedding based question answering}.
\newblock In \emph{Proceedings of the Twelfth {ACM} International Conference on
  Web Search and Data Mining, {WSDM} 2019, Melbourne, VIC, Australia, February
  11-15, 2019}, pages 105--113. {ACM}.

\bibitem[{Hutter(2012)}]{hutter2012human}
Marcus Hutter. 2012.
\newblock The human knowledge compression contest.
\newblock \emph{URL http://prize. hutter1. net}, 6.

\bibitem[{Inan et~al.(2016)Inan, Khosravi, and
  Socher}]{DBLP:journals/corr/InanKS16}
Hakan Inan, Khashayar Khosravi, and Richard Socher. 2016.
\newblock \href {http://arxiv.org/abs/1611.01462} {Tying word vectors and word
  classifiers: {A} loss framework for language modeling}.
\newblock \emph{CoRR}, abs/1611.01462.

\bibitem[{Izacard and Grave(2021)}]{DBLP:conf/eacl/IzacardG21}
Gautier Izacard and Edouard Grave. 2021.
\newblock \href {https://www.aclweb.org/anthology/2021.eacl-main.74/}
  {Leveraging passage retrieval with generative models for open domain question
  answering}.
\newblock In \emph{Proceedings of the 16th Conference of the European Chapter
  of the Association for Computational Linguistics: Main Volume, {EACL} 2021,
  Online, April 19 - 23, 2021}, pages 874--880. Association for Computational
  Linguistics.

\bibitem[{Jelinek(1980)}]{jelinek1980interpolated}
Frederick Jelinek. 1980.
\newblock Interpolated estimation of markov source parameters from sparse data.
\newblock In \emph{Proc. Workshop on Pattern Recognition in Practice, 1980}.

\bibitem[{Ji et~al.(2017)Ji, Tan, Martschat, Choi, and
  Smith}]{DBLP:conf/emnlp/JiTMCS17}
Yangfeng Ji, Chenhao Tan, Sebastian Martschat, Yejin Choi, and Noah~A. Smith.
  2017.
\newblock \href {https://doi.org/10.18653/v1/d17-1195} {Dynamic entity
  representations in neural language models}.
\newblock In \emph{Proceedings of the 2017 Conference on Empirical Methods in
  Natural Language Processing, {EMNLP} 2017, Copenhagen, Denmark, September
  9-11, 2017}, pages 1830--1839. Association for Computational Linguistics.

\bibitem[{Kahneman(2011)}]{kahneman2011}
Daniel Kahneman. 2011.
\newblock \emph{Thinking, Fast and Slow}.
\newblock Farrar, Straus and Giroux.

\bibitem[{Karpukhin et~al.(2020)Karpukhin, Oguz, Min, Lewis, Wu, Edunov, Chen,
  and Yih}]{DBLP:conf/emnlp/KarpukhinOMLWEC20}
Vladimir Karpukhin, Barlas Oguz, Sewon Min, Patrick S.~H. Lewis, Ledell Wu,
  Sergey Edunov, Danqi Chen, and Wen{-}tau Yih. 2020.
\newblock \href {https://doi.org/10.18653/v1/2020.emnlp-main.550} {Dense
  passage retrieval for open-domain question answering}.
\newblock In \emph{Proceedings of the 2020 Conference on Empirical Methods in
  Natural Language Processing, {EMNLP} 2020, Online, November 16-20, 2020},
  pages 6769--6781. Association for Computational Linguistics.

\bibitem[{Khandelwal et~al.(2020{\natexlab{a}})Khandelwal, Fan, Jurafsky,
  Zettlemoyer, and Lewis}]{DBLP:journals/corr/abs-2010-00710}
Urvashi Khandelwal, Angela Fan, Dan Jurafsky, Luke Zettlemoyer, and Mike Lewis.
  2020{\natexlab{a}}.
\newblock \href {http://arxiv.org/abs/2010.00710} {Nearest neighbor machine
  translation}.
\newblock \emph{CoRR}, abs/2010.00710.

\bibitem[{Khandelwal et~al.(2020{\natexlab{b}})Khandelwal, Levy, Jurafsky,
  Zettlemoyer, and Lewis}]{DBLP:conf/iclr/KhandelwalLJZL20}
Urvashi Khandelwal, Omer Levy, Dan Jurafsky, Luke Zettlemoyer, and Mike Lewis.
  2020{\natexlab{b}}.
\newblock \href {https://openreview.net/forum?id=HklBjCEKvH} {Generalization
  through memorization: Nearest neighbor language models}.
\newblock In \emph{8th International Conference on Learning Representations,
  {ICLR} 2020, Addis Ababa, Ethiopia, April 26-30, 2020}. OpenReview.net.

\bibitem[{Kiddon et~al.(2016)Kiddon, Zettlemoyer, and
  Choi}]{DBLP:conf/emnlp/KiddonZC16}
Chlo{\'{e}} Kiddon, Luke Zettlemoyer, and Yejin Choi. 2016.
\newblock \href {https://doi.org/10.18653/v1/d16-1032} {Globally coherent text
  generation with neural checklist models}.
\newblock In \emph{Proceedings of the 2016 Conference on Empirical Methods in
  Natural Language Processing, {EMNLP} 2016, Austin, Texas, USA, November 1-4,
  2016}, pages 329--339. The Association for Computational Linguistics.

\bibitem[{Kingma and Ba(2015)}]{DBLP:journals/corr/KingmaB14}
Diederik~P. Kingma and Jimmy Ba. 2015.
\newblock \href {http://arxiv.org/abs/1412.6980} {Adam: {A} method for
  stochastic optimization}.
\newblock In \emph{3rd International Conference on Learning Representations,
  {ICLR} 2015, San Diego, CA, USA, May 7-9, 2015, Conference Track
  Proceedings}.

\bibitem[{Krause et~al.(2018)Krause, Kahembwe, Murray, and
  Renals}]{DBLP:conf/icml/KrauseK0R18}
Ben Krause, Emmanuel Kahembwe, Iain Murray, and Steve Renals. 2018.
\newblock \href {http://proceedings.mlr.press/v80/krause18a.html} {Dynamic
  evaluation of neural sequence models}.
\newblock In \emph{Proceedings of the 35th International Conference on Machine
  Learning, {ICML} 2018, Stockholmsm{\"{a}}ssan, Stockholm, Sweden, July 10-15,
  2018}, volume~80 of \emph{Proceedings of Machine Learning Research}, pages
  2771--2780. {PMLR}.

\bibitem[{Krause et~al.(2019)Krause, Kahembwe, Murray, and
  Renals}]{DBLP:journals/corr/abs-1904-08378}
Ben Krause, Emmanuel Kahembwe, Iain Murray, and Steve Renals. 2019.
\newblock \href {http://arxiv.org/abs/1904.08378} {Dynamic evaluation of
  transformer language models}.
\newblock \emph{CoRR}, abs/1904.08378.

\bibitem[{Lake and Murphy(2020)}]{DBLP:journals/corr/abs-2008-01766}
Brenden~M. Lake and Gregory~L. Murphy. 2020.
\newblock \href {http://arxiv.org/abs/2008.01766} {Word meaning in minds and
  machines}.
\newblock \emph{CoRR}, abs/2008.01766.

\bibitem[{Lewis et~al.(2020)Lewis, Perez, Piktus, Petroni, Karpukhin, Goyal,
  K{\"{u}}ttler, Lewis, Yih, Rockt{\"{a}}schel, Riedel, and
  Kiela}]{DBLP:conf/nips/LewisPPPKGKLYR020}
Patrick S.~H. Lewis, Ethan Perez, Aleksandra Piktus, Fabio Petroni, Vladimir
  Karpukhin, Naman Goyal, Heinrich K{\"{u}}ttler, Mike Lewis, Wen{-}tau Yih,
  Tim Rockt{\"{a}}schel, Sebastian Riedel, and Douwe Kiela. 2020.
\newblock \href
  {https://proceedings.neurips.cc/paper/2020/hash/6b493230205f780e1bc26945df7481e5-Abstract.html}
  {Retrieval-augmented generation for knowledge-intensive {NLP} tasks}.
\newblock In \emph{Advances in Neural Information Processing Systems 33: Annual
  Conference on Neural Information Processing Systems 2020, NeurIPS 2020,
  December 6-12, 2020, virtual}.

\bibitem[{Lin et~al.(2021)Lin, Hilton, and
  Evans}]{DBLP:journals/corr/abs-2109-07958}
Stephanie Lin, Jacob Hilton, and Owain Evans. 2021.
\newblock \href {http://arxiv.org/abs/2109.07958} {Truthfulqa: Measuring how
  models mimic human falsehoods}.
\newblock \emph{CoRR}, abs/2109.07958.

\bibitem[{Liu et~al.(2019)Liu, Gardner, Belinkov, Peters, and
  Smith}]{liu-etal-2019-linguistic}
Nelson~F. Liu, Matt Gardner, Yonatan Belinkov, Matthew~E. Peters, and Noah~A.
  Smith. 2019.
\newblock \href {https://doi.org/10.18653/v1/N19-1112} {Linguistic knowledge
  and transferability of contextual representations}.
\newblock In \emph{Proceedings of the 2019 Conference of the North {A}merican
  Chapter of the Association for Computational Linguistics: Human Language
  Technologies, Volume 1 (Long and Short Papers)}, pages 1073--1094,
  Minneapolis, Minnesota. Association for Computational Linguistics.

\bibitem[{Liu et~al.(2021{\natexlab{a}})Liu, Yuan, Fu, Jiang, Hayashi, and
  Neubig}]{DBLP:journals/corr/abs-2107-13586}
Pengfei Liu, Weizhe Yuan, Jinlan Fu, Zhengbao Jiang, Hiroaki Hayashi, and
  Graham Neubig. 2021{\natexlab{a}}.
\newblock \href {http://arxiv.org/abs/2107.13586} {Pre-train, prompt, and
  predict: {A} systematic survey of prompting methods in natural language
  processing}.
\newblock \emph{CoRR}, abs/2107.13586.

\bibitem[{Liu et~al.(2021{\natexlab{b}})Liu, Yu, Rimell, and
  Blunsom}]{liu-etal-2021-pretraining}
Qi~Liu, Lei Yu, Laura Rimell, and Phil Blunsom. 2021{\natexlab{b}}.
\newblock \href {https://doi.org/10.1162/tacl_a_00390} {Pretraining the noisy
  channel model for task-oriented dialogue}.
\newblock \emph{Transactions of the Association for Computational Linguistics},
  9:657--674.

\bibitem[{Liu et~al.(2020)Liu, Zhou, Zhao, Wang, Ju, Deng, and
  Wang}]{DBLP:conf/aaai/LiuZ0WJD020}
Weijie Liu, Peng Zhou, Zhe Zhao, Zhiruo Wang, Qi~Ju, Haotang Deng, and Ping
  Wang. 2020.
\newblock \href {https://aaai.org/ojs/index.php/AAAI/article/view/5681}
  {{K-BERT:} enabling language representation with knowledge graph}.
\newblock In \emph{The Thirty-Fourth {AAAI} Conference on Artificial
  Intelligence, {AAAI} 2020, The Thirty-Second Innovative Applications of
  Artificial Intelligence Conference, {IAAI} 2020, The Tenth {AAAI} Symposium
  on Educational Advances in Artificial Intelligence, {EAAI} 2020, New York,
  NY, USA, February 7-12, 2020}, pages 2901--2908. {AAAI} Press.

\bibitem[{Logan et~al.(2019)Logan, Liu, Peters, Gardner, and
  Singh}]{DBLP:conf/acl/LoganLPGS19}
Robert~L. Logan, Nelson~F. Liu, Matthew~E. Peters, Matt Gardner, and Sameer
  Singh. 2019.
\newblock \href {https://doi.org/10.18653/v1/p19-1598} {Barack's wife hillary:
  Using knowledge graphs for fact-aware language modeling}.
\newblock In \emph{Proceedings of the 57th Conference of the Association for
  Computational Linguistics, {ACL} 2019, Florence, Italy, July 28- August 2,
  2019, Volume 1: Long Papers}, pages 5962--5971. Association for Computational
  Linguistics.

\bibitem[{de~Masson~d'Autume et~al.(2019)de~Masson~d'Autume, Ruder, Kong, and
  Yogatama}]{mbpa2019}
Cyprien de~Masson~d'Autume, Sebastian Ruder, Lingpeng Kong, and Dani Yogatama.
  2019.
\newblock Episodic memory in lifelong language learning.
\newblock In \emph{Advances in Neural Information Processing Systems}.

\bibitem[{Merity et~al.(2017)Merity, Xiong, Bradbury, and
  Socher}]{DBLP:conf/iclr/MerityX0S17}
Stephen Merity, Caiming Xiong, James Bradbury, and Richard Socher. 2017.
\newblock \href {https://openreview.net/forum?id=Byj72udxe} {Pointer sentinel
  mixture models}.
\newblock In \emph{5th International Conference on Learning Representations,
  {ICLR} 2017, Toulon, France, April 24-26, 2017, Conference Track
  Proceedings}. OpenReview.net.

\bibitem[{Minervini et~al.(2020)Minervini, Bosnjak, Rockt{\"{a}}schel, Riedel,
  and Grefenstette}]{DBLP:conf/aaai/MinerviniBR0G20}
Pasquale Minervini, Matko Bosnjak, Tim Rockt{\"{a}}schel, Sebastian Riedel, and
  Edward Grefenstette. 2020.
\newblock \href {https://aaai.org/ojs/index.php/AAAI/article/view/5962}
  {Differentiable reasoning on large knowledge bases and natural language}.
\newblock In \emph{The Thirty-Fourth {AAAI} Conference on Artificial
  Intelligence, {AAAI} 2020, The Thirty-Second Innovative Applications of
  Artificial Intelligence Conference, {IAAI} 2020, The Tenth {AAAI} Symposium
  on Educational Advances in Artificial Intelligence, {EAAI} 2020, New York,
  NY, USA, February 7-12, 2020}, pages 5182--5190. {AAAI} Press.

\bibitem[{Moon et~al.(2019)Moon, Shah, Kumar, and Subba}]{Moon2019opendialkg}
Seungwhan Moon, Pararth Shah, Anuj Kumar, and Rajen Subba. 2019.
\newblock Opendialkg: Explainable conversational reasoning with attention-based
  walks over knowledge graphs.
\newblock In \emph{Proceedings of the 57th Annual Meeting of the Association
  for Computational Linguistics}.

\bibitem[{Nadeau and Sekine(2007)}]{nadeau2007survey}
David Nadeau and Satoshi Sekine. 2007.
\newblock A survey of named entity recognition and classification.
\newblock \emph{Lingvisticae Investigationes}, 30(1):3--26.

\bibitem[{Nye et~al.(2021)Nye, Tessler, Tenenbaum, and Lake}]{nye2021}
Maxwell Nye, Michael~Henry Tessler, Joshua~B. Tenenbaum, and Brenden~M. Lake.
  2021.
\newblock \href {http://arxiv.org/abs/2107.02794} {Improving coherence and
  consistency in neural sequence models with dual-system, neuro-symbolic
  reasoning}.
\newblock \emph{CoRR}, abs/2107.02794.

\bibitem[{Ostendorff et~al.(2019)Ostendorff, Bourgonje, Berger, Schneider,
  Rehm, and Gipp}]{DBLP:conf/konvens/OstendorffBBSRG19}
Malte Ostendorff, Peter Bourgonje, Maria Berger, Juli{\'{a}}n~Moreno Schneider,
  Georg Rehm, and Bela Gipp. 2019.
\newblock \href
  {https://corpora.linguistik.uni-erlangen.de/data/konvens/proceedings/papers/germeval/Germeval\_Task1\_paper\_3.pdf}
  {Enriching {BERT} with knowledge graph embeddings for document
  classification}.
\newblock In \emph{Proceedings of the 15th Conference on Natural Language
  Processing, {KONVENS} 2019, Erlangen, Germany, October 9-11, 2019}.

\bibitem[{Parvez et~al.(2018)Parvez, Chakraborty, Ray, and
  Chang}]{DBLP:conf/acl/ChangPCR18}
Md.~Rizwan Parvez, Saikat Chakraborty, Baishakhi Ray, and Kai{-}Wei Chang.
  2018.
\newblock \href {https://doi.org/10.18653/v1/P18-1221} {Building language
  models for text with named entities}.
\newblock In \emph{Proceedings of the 56th Annual Meeting of the Association
  for Computational Linguistics, {ACL} 2018, Melbourne, Australia, July 15-20,
  2018, Volume 1: Long Papers}, pages 2373--2383. Association for Computational
  Linguistics.

\bibitem[{Peters et~al.(2019)Peters, Neumann, IV, Schwartz, Joshi, Singh, and
  Smith}]{DBLP:conf/emnlp/PetersNLSJSS19}
Matthew~E. Peters, Mark Neumann, Robert L.~Logan IV, Roy Schwartz, Vidur Joshi,
  Sameer Singh, and Noah~A. Smith. 2019.
\newblock \href {https://doi.org/10.18653/v1/D19-1005} {Knowledge enhanced
  contextual word representations}.
\newblock In \emph{Proceedings of the 2019 Conference on Empirical Methods in
  Natural Language Processing and the 9th International Joint Conference on
  Natural Language Processing, {EMNLP-IJCNLP} 2019, Hong Kong, China, November
  3-7, 2019}, pages 43--54. Association for Computational Linguistics.

\bibitem[{Petroni et~al.(2019)Petroni, Rockt{\"{a}}schel, Riedel, Lewis,
  Bakhtin, Wu, and Miller}]{DBLP:conf/emnlp/PetroniRRLBWM19}
Fabio Petroni, Tim Rockt{\"{a}}schel, Sebastian Riedel, Patrick S.~H. Lewis,
  Anton Bakhtin, Yuxiang Wu, and Alexander~H. Miller. 2019.
\newblock \href {https://doi.org/10.18653/v1/D19-1250} {Language models as
  knowledge bases?}
\newblock In \emph{Proceedings of the 2019 Conference on Empirical Methods in
  Natural Language Processing and the 9th International Joint Conference on
  Natural Language Processing, {EMNLP-IJCNLP} 2019, Hong Kong, China, November
  3-7, 2019}, pages 2463--2473. Association for Computational Linguistics.

\bibitem[{Radford et~al.(2019)Radford, Wu, Child, Luan, Amodei, and
  Sutskever}]{radford2019language}
Alec Radford, Jeffrey Wu, Rewon Child, David Luan, Dario Amodei, and Ilya
  Sutskever. 2019.
\newblock Language models are unsupervised multitask learners.
\newblock \emph{OpenAI blog}, 1(8):9.

\bibitem[{Ramos et~al.(2003)}]{ramos2003using}
Juan Ramos et~al. 2003.
\newblock Using tf-idf to determine word relevance in document queries.
\newblock In \emph{Proceedings of the first instructional conference on machine
  learning}, volume 242, pages 29--48. Citeseer.

\bibitem[{Ratinov and Roth(2009)}]{ratinov2009design}
Lev Ratinov and Dan Roth. 2009.
\newblock Design challenges and misconceptions in named entity recognition.
\newblock In \emph{Proceedings of the Thirteenth Conference on Computational
  Natural Language Learning (CoNLL-2009)}, pages 147--155.

\bibitem[{Rebele et~al.(2016)Rebele, Suchanek, Hoffart, Biega, Kuzey, and
  Weikum}]{DBLP:conf/semweb/RebeleSHBKW16}
Thomas Rebele, Fabian~M. Suchanek, Johannes Hoffart, Joanna Biega, Erdal Kuzey,
  and Gerhard Weikum. 2016.
\newblock \href {https://doi.org/10.1007/978-3-319-46547-0\_19} {{YAGO:} {A}
  multilingual knowledge base from wikipedia, wordnet, and geonames}.
\newblock In \emph{The Semantic Web - {ISWC} 2016 - 15th International Semantic
  Web Conference, Kobe, Japan, October 17-21, 2016, Proceedings, Part {II}},
  volume 9982 of \emph{Lecture Notes in Computer Science}, pages 177--185.

\bibitem[{Schlichtkrull et~al.(2018)Schlichtkrull, Kipf, Bloem, van~den Berg,
  Titov, and Welling}]{DBLP:conf/esws/SchlichtkrullKB18}
Michael~Sejr Schlichtkrull, Thomas~N. Kipf, Peter Bloem, Rianne van~den Berg,
  Ivan Titov, and Max Welling. 2018.
\newblock \href {https://doi.org/10.1007/978-3-319-93417-4\_38} {Modeling
  relational data with graph convolutional networks}.
\newblock In \emph{The Semantic Web - 15th International Conference, {ESWC}
  2018, Heraklion, Crete, Greece, June 3-7, 2018, Proceedings}, volume 10843 of
  \emph{Lecture Notes in Computer Science}, pages 593--607. Springer.

\bibitem[{Srivastava et~al.(2014)Srivastava, Hinton, Krizhevsky, Sutskever, and
  Salakhutdinov}]{DBLP:journals/jmlr/SrivastavaHKSS14}
Nitish Srivastava, Geoffrey~E. Hinton, Alex Krizhevsky, Ilya Sutskever, and
  Ruslan Salakhutdinov. 2014.
\newblock \href {http://dl.acm.org/citation.cfm?id=2670313} {Dropout: a simple
  way to prevent neural networks from overfitting}.
\newblock \emph{J. Mach. Learn. Res.}, 15(1):1929--1958.

\bibitem[{Sun et~al.(2019)Sun, Wang, Li, Feng, Chen, Zhang, Tian, Zhu, Tian,
  and Wu}]{DBLP:journals/corr/abs-1904-09223}
Yu~Sun, Shuohuan Wang, Yu{-}Kun Li, Shikun Feng, Xuyi Chen, Han Zhang, Xin
  Tian, Danxiang Zhu, Hao Tian, and Hua Wu. 2019.
\newblock \href {http://arxiv.org/abs/1904.09223} {{ERNIE:} enhanced
  representation through knowledge integration}.
\newblock \emph{CoRR}, abs/1904.09223.

\bibitem[{Thulke et~al.(2021)Thulke, Daheim, Dugast, and
  Ney}]{thulke2021efficient}
David Thulke, Nico Daheim, Christian Dugast, and Hermann Ney. 2021.
\newblock Efficient retrieval augmented generation from unstructured knowledge
  for task-oriented dialog.
\newblock \emph{arXiv preprint arXiv:2102.04643}.

\bibitem[{Trouillon et~al.(2016)Trouillon, Welbl, Riedel, Gaussier, and
  Bouchard}]{DBLP:conf/icml/TrouillonWRGB16}
Th{\'{e}}o Trouillon, Johannes Welbl, Sebastian Riedel, {\'{E}}ric Gaussier,
  and Guillaume Bouchard. 2016.
\newblock \href {http://proceedings.mlr.press/v48/trouillon16.html} {Complex
  embeddings for simple link prediction}.
\newblock In \emph{Proceedings of the 33nd International Conference on Machine
  Learning, {ICML} 2016, New York City, NY, USA, June 19-24, 2016}, volume~48
  of \emph{{JMLR} Workshop and Conference Proceedings}, pages 2071--2080.
  JMLR.org.

\bibitem[{Vaswani et~al.(2017)Vaswani, Shazeer, Parmar, Uszkoreit, Jones,
  Gomez, Kaiser, and Polosukhin}]{DBLP:conf/nips/VaswaniSPUJGKP17}
Ashish Vaswani, Noam Shazeer, Niki Parmar, Jakob Uszkoreit, Llion Jones,
  Aidan~N. Gomez, Lukasz Kaiser, and Illia Polosukhin. 2017.
\newblock \href
  {https://proceedings.neurips.cc/paper/2017/hash/3f5ee243547dee91fbd053c1c4a845aa-Abstract.html}
  {Attention is all you need}.
\newblock In \emph{Advances in Neural Information Processing Systems 30: Annual
  Conference on Neural Information Processing Systems 2017, December 4-9, 2017,
  Long Beach, CA, {USA}}, pages 5998--6008.

\bibitem[{Verga et~al.(2021)Verga, Sun, Soares, and
  Cohen}]{DBLP:conf/naacl/VergaSSC21}
Pat Verga, Haitian Sun, Livio~Baldini Soares, and William~W. Cohen. 2021.
\newblock \href {https://doi.org/10.18653/v1/2021.naacl-main.288} {Adaptable
  and interpretable neural memoryover symbolic knowledge}.
\newblock In \emph{Proceedings of the 2021 Conference of the North American
  Chapter of the Association for Computational Linguistics: Human Language
  Technologies, {NAACL-HLT} 2021, Online, June 6-11, 2021}, pages 3678--3691.
  Association for Computational Linguistics.

\bibitem[{Wang et~al.(2020)Wang, Liu, and
  Song}]{DBLP:journals/corr/abs-2010-11967}
Chenguang Wang, Xiao Liu, and Dawn Song. 2020.
\newblock \href {http://arxiv.org/abs/2010.11967} {Language models are open
  knowledge graphs}.
\newblock \emph{CoRR}, abs/2010.11967.

\bibitem[{Wang et~al.(2018{\natexlab{a}})Wang, Zhang, Xie, and
  Guo}]{DBLP:conf/www/WangZXG18}
Hongwei Wang, Fuzheng Zhang, Xing Xie, and Minyi Guo. 2018{\natexlab{a}}.
\newblock \href {https://doi.org/10.1145/3178876.3186175} {{DKN:} deep
  knowledge-aware network for news recommendation}.
\newblock In \emph{Proceedings of the 2018 World Wide Web Conference on World
  Wide Web, {WWW} 2018, Lyon, France, April 23-27, 2018}, pages 1835--1844.
  {ACM}.

\bibitem[{Wang et~al.(2019)Wang, Zhang, Zhao, Li, Xie, and
  Guo}]{DBLP:conf/www/WangZZLXG19}
Hongwei Wang, Fuzheng Zhang, Miao Zhao, Wenjie Li, Xing Xie, and Minyi Guo.
  2019.
\newblock \href {https://doi.org/10.1145/3308558.3313411} {Multi-task feature
  learning for knowledge graph enhanced recommendation}.
\newblock In \emph{The World Wide Web Conference, {WWW} 2019, San Francisco,
  CA, USA, May 13-17, 2019}, pages 2000--2010. {ACM}.

\bibitem[{Wang et~al.(2021{\natexlab{a}})Wang, Li, Aslan, and
  Vinyals}]{wang-etal-2021-wikigraphs}
Luyu Wang, Yujia Li, Ozlem Aslan, and Oriol Vinyals. 2021{\natexlab{a}}.
\newblock \href {https://www.aclweb.org/anthology/2021.textgraphs-1.7}
  {{W}iki{G}raphs: A {W}ikipedia text - knowledge graph paired dataset}.
\newblock In \emph{Proceedings of the Fifteenth Workshop on Graph-Based Methods
  for Natural Language Processing (TextGraphs-15)}, pages 67--82, Mexico City,
  Mexico. Association for Computational Linguistics.

\bibitem[{Wang et~al.(2018{\natexlab{b}})Wang, Pan, Huang, Zhang, Jiang, Ji,
  and Knight}]{DBLP:conf/inlg/0005PHZJJK18}
Qingyun Wang, Xiaoman Pan, Lifu Huang, Boliang Zhang, Zhiying Jiang, Heng Ji,
  and Kevin Knight. 2018{\natexlab{b}}.
\newblock \href {https://doi.org/10.18653/v1/w18-6502} {Describing a knowledge
  base}.
\newblock In \emph{Proceedings of the 11th International Conference on Natural
  Language Generation, Tilburg University, The Netherlands, November 5-8,
  2018}, pages 10--21. Association for Computational Linguistics.

\bibitem[{Wang et~al.(2021{\natexlab{b}})Wang, Gao, Zhu, Zhang, Liu, Li, and
  Tang}]{wang2021kepler}
Xiaozhi Wang, Tianyu Gao, Zhaocheng Zhu, Zhengyan Zhang, Zhiyuan Liu, Juanzi
  Li, and Jian Tang. 2021{\natexlab{b}}.
\newblock Kepler: A unified model for knowledge embedding and pre-trained
  language representation.
\newblock \emph{Transactions of the Association for Computational Linguistics},
  9:176--194.

\bibitem[{Yang and Mitchell(2019)}]{DBLP:journals/corr/abs-1902-09091}
Bishan Yang and Tom~M. Mitchell. 2019.
\newblock \href {http://arxiv.org/abs/1902.09091} {Leveraging knowledge bases
  in lstms for improving machine reading}.
\newblock \emph{CoRR}, abs/1902.09091.

\bibitem[{Yang et~al.(2017)Yang, Blunsom, Dyer, and
  Ling}]{DBLP:conf/emnlp/YangBDL17}
Zichao Yang, Phil Blunsom, Chris Dyer, and Wang Ling. 2017.
\newblock \href {https://doi.org/10.18653/v1/d17-1197} {Reference-aware
  language models}.
\newblock In \emph{Proceedings of the 2017 Conference on Empirical Methods in
  Natural Language Processing, {EMNLP} 2017, Copenhagen, Denmark, September
  9-11, 2017}, pages 1850--1859. Association for Computational Linguistics.

\bibitem[{Yao et~al.(2019)Yao, Mao, and
  Luo}]{DBLP:journals/corr/abs-1909-03193}
Liang Yao, Chengsheng Mao, and Yuan Luo. 2019.
\newblock \href {http://arxiv.org/abs/1909.03193} {{KG-BERT:} {BERT} for
  knowledge graph completion}.
\newblock \emph{CoRR}, abs/1909.03193.

\bibitem[{Yasunaga et~al.(2021)Yasunaga, Ren, Bosselut, Liang, and
  Leskovec}]{DBLP:journals/corr/abs-2104-06378}
Michihiro Yasunaga, Hongyu Ren, Antoine Bosselut, Percy Liang, and Jure
  Leskovec. 2021.
\newblock \href {http://arxiv.org/abs/2104.06378} {{QA-GNN:} reasoning with
  language models and knowledge graphs for question answering}.
\newblock \emph{CoRR}, abs/2104.06378.

\bibitem[{Yogatama et~al.(2021)Yogatama, de~Masson~d'Autume, and
  Kong}]{DBLP:journals/tacl/YogatamadK21}
Dani Yogatama, Cyprien de~Masson~d'Autume, and Lingpeng Kong. 2021.
\newblock \href {https://transacl.org/ojs/index.php/tacl/article/view/2693}
  {Adaptive semiparametric language models}.
\newblock \emph{Trans. Assoc. Comput. Linguistics}, 9:362--373.

\bibitem[{Zhang et~al.(2016)Zhang, Yuan, Lian, Xie, and
  Ma}]{DBLP:conf/kdd/ZhangYLXM16}
Fuzheng Zhang, Nicholas~Jing Yuan, Defu Lian, Xing Xie, and Wei{-}Ying Ma.
  2016.
\newblock \href {https://doi.org/10.1145/2939672.2939673} {Collaborative
  knowledge base embedding for recommender systems}.
\newblock In \emph{Proceedings of the 22nd {ACM} {SIGKDD} International
  Conference on Knowledge Discovery and Data Mining, San Francisco, CA, USA,
  August 13-17, 2016}, pages 353--362. {ACM}.

\bibitem[{Zhang and Chen(2018)}]{DBLP:conf/nips/ZhangC18}
Muhan Zhang and Yixin Chen. 2018.
\newblock \href
  {https://proceedings.neurips.cc/paper/2018/hash/53f0d7c537d99b3824f0f99d62ea2428-Abstract.html}
  {Link prediction based on graph neural networks}.
\newblock In \emph{Advances in Neural Information Processing Systems 31: Annual
  Conference on Neural Information Processing Systems 2018, NeurIPS 2018,
  December 3-8, 2018, Montr{\'{e}}al, Canada}, pages 5171--5181.

\bibitem[{Zhang et~al.(2019)Zhang, Tay, Yao, and
  Liu}]{DBLP:conf/nips/0007TYL19}
Shuai Zhang, Yi~Tay, Lina Yao, and Qi~Liu. 2019.
\newblock \href
  {https://proceedings.neurips.cc/paper/2019/hash/d961e9f236177d65d21100592edb0769-Abstract.html}
  {Quaternion knowledge graph embeddings}.
\newblock In \emph{Advances in Neural Information Processing Systems 32: Annual
  Conference on Neural Information Processing Systems 2019, NeurIPS 2019,
  December 8-14, 2019, Vancouver, BC, Canada}, pages 2731--2741.

\bibitem[{Zhang et~al.(2018)Zhang, Dai, Kozareva, Smola, and
  Song}]{DBLP:conf/aaai/ZhangDKSS18}
Yuyu Zhang, Hanjun Dai, Zornitsa Kozareva, Alexander~J. Smola, and Le~Song.
  2018.
\newblock \href
  {https://www.aaai.org/ocs/index.php/AAAI/AAAI18/paper/view/16983}
  {Variational reasoning for question answering with knowledge graph}.
\newblock In \emph{Proceedings of the Thirty-Second {AAAI} Conference on
  Artificial Intelligence, (AAAI-18), the 30th innovative Applications of
  Artificial Intelligence (IAAI-18), and the 8th {AAAI} Symposium on
  Educational Advances in Artificial Intelligence (EAAI-18), New Orleans,
  Louisiana, USA, February 2-7, 2018}, pages 6069--6076. {AAAI} Press.

\bibitem[{Zhou et~al.(2018)Zhou, Li, Dong, Liu, Chen, Zhao, Yu, and
  Wu}]{DBLP:conf/acl/WuLCZDYZL18}
Xiangyang Zhou, Lu~Li, Daxiang Dong, Yi~Liu, Ying Chen, Wayne~Xin Zhao, Dianhai
  Yu, and Hua Wu. 2018.
\newblock \href {https://doi.org/10.18653/v1/P18-1103} {Multi-turn response
  selection for chatbots with deep attention matching network}.
\newblock In \emph{Proceedings of the 56th Annual Meeting of the Association
  for Computational Linguistics, {ACL} 2018, Melbourne, Australia, July 15-20,
  2018, Volume 1: Long Papers}, pages 1118--1127. Association for Computational
  Linguistics.

\end{thebibliography}
\bibliographystyle{acl_natbib}

\end{document}